\documentclass[11pt]{article}

\usepackage[final]{acl}

\usepackage{times}
\usepackage{latexsym}

\usepackage[T1]{fontenc}

\usepackage[utf8]{inputenc}

\usepackage{microtype}

\usepackage{inconsolata}

\usepackage{graphicx}
\usepackage{hyperref}
\usepackage{url}
\usepackage[utf8]{inputenc} 
\usepackage[T1]{fontenc}    
\usepackage{hyperref}       
\usepackage{url}            
\usepackage{booktabs}       
\usepackage{amsfonts}       
\usepackage{nicefrac}       
\usepackage{microtype}      
\usepackage{xcolor}         

\usepackage{multirow}
\usepackage{graphicx}
\usepackage{amssymb}
\usepackage{mathtools}
\usepackage{amsthm}
\usepackage{multirow}
\usepackage{colortbl}
\usepackage{pifont}
\usepackage{subfigure}
\usepackage{xspace}
\usepackage{xcolor}          
\usepackage{todonotes}
\usepackage{textcomp}
\usepackage{caption}
\usepackage{wrapfig}
\usepackage{float}
\usepackage{textcomp}
\usepackage{pifont}
\usepackage{hyperref}

\usepackage{longtable}
\usepackage{dashrule}
\usepackage{makecell}
\usepackage{threeparttable}
\usepackage{enumitem}
\usepackage[normalem]{ulem}
\usepackage{tabularx}
\usepackage{subcaption}

\title{TabularMath: Understanding Math Reasoning over Tables \\with Large Language Models}

\newcommand{\cmark}{{\color{green!40!black}\ding{51}}} 
\newcommand{\xmark}{{\color{red!40!black}\ding{55}}}   

\newcommand{\benchmark}{\text{TabularMath}\xspace}
\newcommand{\Pipeline}{\textsc{AutoT2T}\xspace}

\usepackage[most]{tcolorbox}
\newtcolorbox[auto counter, number freestyle={\noexpand\arabic{\tcbcounter}}]{definedfindingbox}[2][]{%
    enhanced,
colback=blue!5!white,
colframe=blue!60!black,
colbacktitle=blue!15!white,
coltitle=black,
    title=#2,
    #1
}

\newtcolorbox[auto counter, number freestyle={\noexpand\arabic{\tcbcounter}}]{definedbox}[2][]{%
    enhanced,
    colback=black!5!white,
    colframe=black!75!white,
    title=#2,
    #1
}

\newtcolorbox{formalizedprompt}[1][]{%
  enhanced,
  breakable,
  colback=gray!5,
  colframe=gray!80,
  fonttitle=\bfseries,
  title=Formalize2Tabular\_prompt,
  coltitle=black,
  sharp corners,
  boxrule=0.8pt,
  arc=2mm,
  left=4pt,
  right=4pt,
  top=6pt,
  bottom=6pt,
  width=\textwidth,
  #1
}

\author{
  Shi-Yu Tian$^{1,2 *}$, Zhi Zhou$^{1} \thanks{Equal contribution.}$, Wei Dong$^{2 *}$, Kun-Yang Yu$^{1,2}$, Ming Yang$^{1,2}$, Zi-Jian Cheng$^{1,3}$, \\ 
  \textbf{Lan-Zhe Guo$^{1,3 \dagger}$, Yu-Feng Li$^{1,2}\thanks{Corresponding author.}$} \\
  $^1${National Key Laboratory for Novel Software Technology, Nanjing University}\\
  $^2${School of Artificial Intelligence, Nanjing University} \\
  $^3${School of Intelligence Science and Technology, Nanjing University} \\
  \texttt{\{tiansy,zhouz,guolz,liyf\}@lamda.nju.edu.cn}
}

\begin{document}
\maketitle
\begin{abstract}
Mathematical reasoning has long been a key benchmark for evaluating large language models (LLMs). 
Although substantial progress has been made on math word problems, the need for reasoning over tabular data in real-world applications has been overlooked. 
For instance, applications such as business intelligence demand not only multi-step numerical reasoning with tables but also robustness to incomplete or inconsistent information. 
However, comprehensive evaluation in this area is severely limited, constrained by the reliance on manually collected tables that are difficult to scale and the lack of coverage for potential traps encountered in real-world scenarios.
To address this problem, we propose \Pipeline, a neuro-symbolic framework that controllably transforms math word problems into scalable and verified tabular reasoning tasks, enabling the evaluation of both accuracy and robustness.
Building on this pipeline, we develop \benchmark, a benchmark comprising three progressively complex subsets and an imperfect subset, with their corresponding image version.
Our study reveals three key observations: 
(1) Table complexity and reasoning difficulty impact reasoning performance jointly; 
(2) Low-quality tables pose severe risks to reliable reasoning in current LLMs; 
(3) Different table modalities show similar trends, with text-based tables typically being easier for models to reason over even for MLLMs.
In-depth analyses are conducted for each observation to guide future research.
\end{abstract}

\section{Introduction}

Mathematical reasoning has long been a critical benchmark for evaluating the capabilities of large language models (LLMs). The field has advanced remarkably in recent years~\citep{GPT4,guo2025deepseekr1}, with many single-scenario benchmarks now considered largely solved~\citep{hosseini2014AddSub,patel2021svamp,Cobbe2021gsm8k}. This progress has prompted a shift in research focus toward real-world applications, particularly reasoning over semi-structured data like tables~\citep{lu2022tabmwp}. Unlike plain text, tables present information in a highly structured and organized format, making them indispensable in domains such as business intelligence~\citep{zhang2024tablellm} and financial forecasting~\citep{zhu2021tatqa}. 

Nevertheless, real-world table reasoning scenarios present significant challenges for LLMs. For example, in the financial sector, the need to process large-scale tables continues to grow with the increasing volume and complexity of data, alongside stricter requirements for reliability and security~\citep{bradley2024finaticaltable2,zavitsanos2024finantialtable}. 
In quarterly financial reports, models are expected not only to perform cross-column computations on numerous metrics like revenue, profit, and liabilities but also to verify numerical consistency (e.g., ensuring total assets equal the sum of liabilities and equity). Failure to properly interpret the data or detect inconsistencies can lead to 
severe consequences in downstream applications like investment decisions and risk assessment~
\citep{cerchiello2016finaltialrisk}.

\begin{figure}
    \centering
    \includegraphics[width=0.4\textwidth]{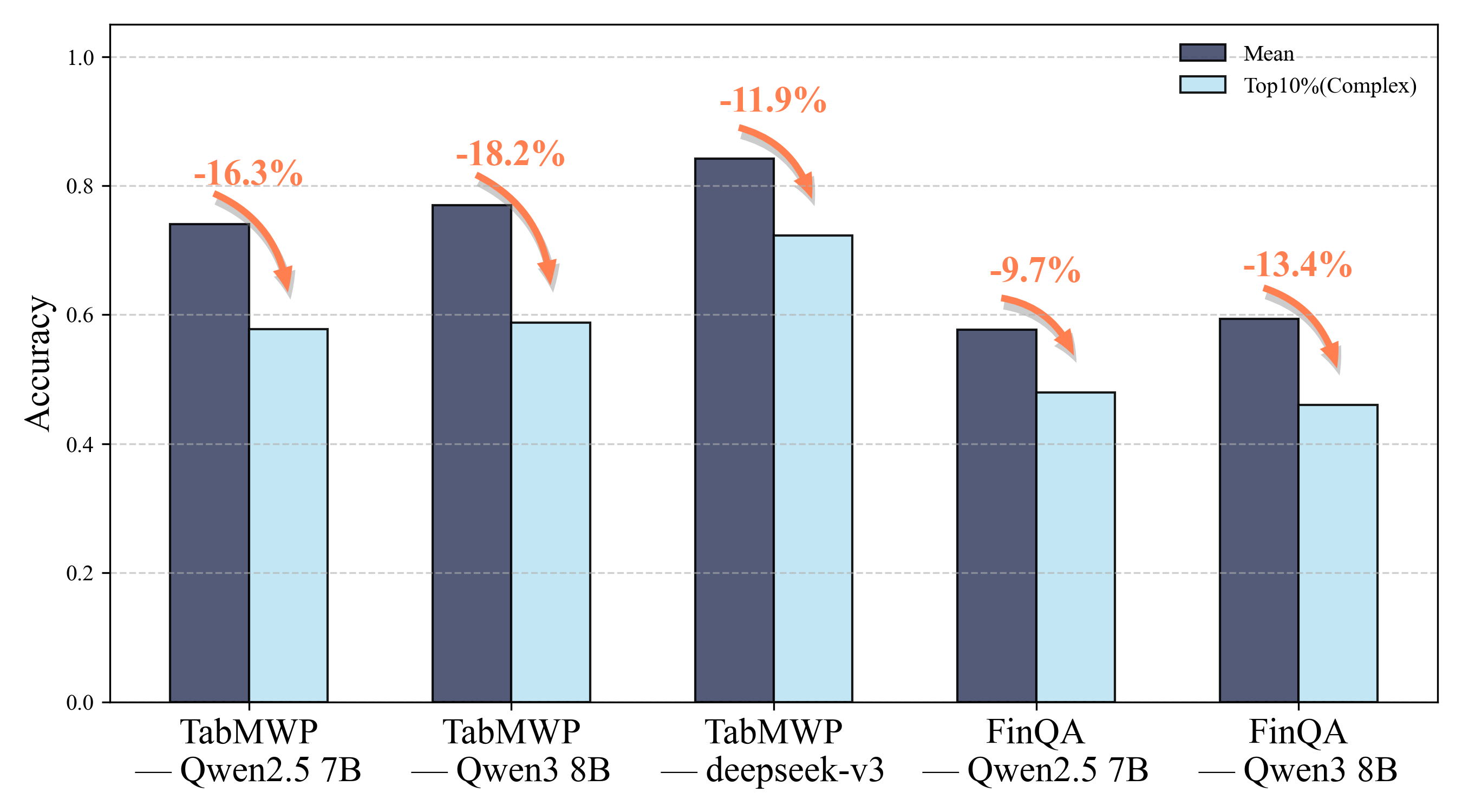}
    \vspace{-0.5em}
    \caption{Model performance comparison between average questions and top10\% complex questions.}
    \label{fig:intro_performace}
    \vspace{-1.5em}
\end{figure}

Despite prior works~\citep{zhu2021tatqa,chen2021finqa,lu2022tabmwp} addressing some aspects of tabular mathematical reasoning, these efforts have been limited in table scale and primarily focused on accuracy of perfectly crafted problems.
Specifically, existing tabular benchmarks largely rely on manual annotation and collection, making it difficult to scale the datasets effectively. 
As a result, these benchmarks fail to explore the limits of LLMs' reasoning capabilities on more complex tables, where the models perform worse (as shown in Fig.~\ref{fig:intro_performace}). 
Then, current benchmarks have not adequately assessed the robustness of tabular mathematical reasoning, overlooking the risk of LLMs providing hallucinated answers when faced with incomplete and inconsistent data. 
Therefore, to systematically assess model capabilities across multiple dimensions, a comprehensive benchmark is crucial. In this context, there is an urgent need for a more complete and systematic evaluation framework to thoroughly explore and challenge the boundaries of existing models.

To address the above limitations, we propose an \textbf{Auto}matic \textbf{T}ext-\textbf{t}o-\textbf{T}able generation framework, \Pipeline. It is a neuro-symbolic pipeline that converts math word problems into scalable and verified tabular reasoning tasks without human annotation, enabling the evaluation of both
accuracy and robustness. To facilitate standardized evaluation and fair comparison, we construct a comprehensive tabular math reasoning benchmark \textbf{\benchmark} based on \Pipeline. It includes three progressively difficult subsets (\emph{Easy, Medium, Hard}) as well as an \emph{Imperfect} subset aimed at evaluating the robustness of models in the face of incomplete or inconsistent tabular data, covering both table complexity and robustness dimensions. Based on this, we conduct systematic experiments and analyses on 18 open-source and proprietary models. The results are organized around the following three research questions and lead to several key observations.


\begin{enumerate}[itemsep=0.2em, leftmargin=2em]
    \item \textbf{\textit{How does table complexity affect mathematical reasoning?} Tabular complexity and reasoning difficulty impact reasoning performance jointly.} Nearly all models suffer significant performance drops when transitioning from pure text to tabular modalities, with degradation increasing as table complexity grows. The coupling between retrieval and reasoning forms a core bottleneck, with pure retrieval being substantially easier than joint reasoning and retrieval (performance gap exceeding 20\% on average).
    \item \textbf{\textit{How does table quality affect mathematical reasoning?} Low-quality tables pose severe risks to reliable reasoning in current LLMs.}. When tables contain missing or contradictory information, most models fail to identify these flaws and produce misleading answers, with error rates exceeding 50\% in some cases. Moreover, when models are informed that inputs may contain imperfect expression, they experience performance degradation on well-defined problems, demonstrating a trade-off between solvency and discriminative ability.
    \item \textbf{\textit{How does table representation affect mathematical reasoning?} 
    Different table modalities show similar trends, with text-based tables typically being easier for models to reason over.}
    Across models and difficulty levels, image- and text-based tables show similar trends, while even multimodal models achieve comparable or higher accuracy on text-based tables. Among textual formats, key-value structured formats such as JSON and serialization perform better.
\end{enumerate}

Overall, we conduct a systematic and in-depth analysis of tabular mathematical reasoning from the perspectives of table complexity, table quality, and table representation, complemented by additional discussions. This work represents an exploratory step toward multimodal reasoning over structured data, laying the groundwork for addressing these challenges in future research.


\section{Related work}

\begin{figure*}
    \centering
    \includegraphics[width=0.9\linewidth]{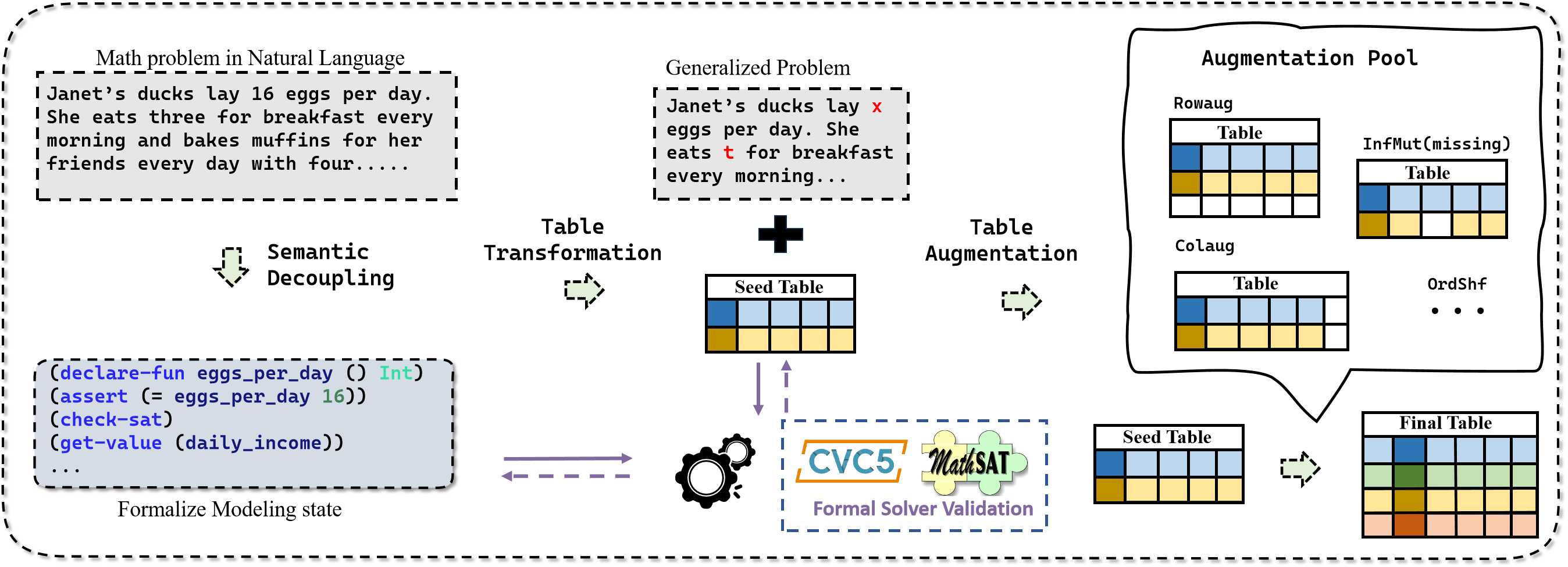}
    \caption{An overview of \Pipeline pipeline.}
    \label{fig: overview pipeline}
    \vspace{-1em}
\end{figure*}

\textbf{Math Reasoning and Benchmark Evaluation.}
Mathematical reasoning serves as a key benchmark for evaluating the capabilities of large language models (LLMs) due to its verifiable nature. Early progress was made on elementary-level math problems using datasets such as GSM8K~\citep{Cobbe2021gsm8k}, MultiArith~\citep{koncel2016multiarith}, and SVAMP~\citep{patel2021svamp}, where methods like in-context learning~\citep{wei2022cot,gao2023pal}, supervised fine-tuning~\citep{li2024neurosymbolic}, and reinforcement learning~\citep{guo2025deepseekr1} demonstrated strong performance. Since then, researchers have questioned the accuracy of current assessments of large models’ mathematical reasoning, exploring approaches such as neural-symbolic methods~\citep{mirzadeh2024gsmsymbolic}. These neural symbolic methods are also widely used in multimodal benchmark generation.~\citep{zhou2024decoop,shang2026maptab,yang2026nesyroute,pyma2026thinkingblueprintsassistingvisionlanguage,pyhuang2026semanticspaceexplorationexploitationrlvr}. A growing area of interest is the robustness of mathematical reasoning~\citep{zhou2024mathattack,shi2023gsmic}, specifically, whether models can refrain from generating hallucinations when faced with incomplete or logically deceptive prompts~\citep{tian2025vcsearch,zhao2024mathtrap}. These types of descriptive verification issues are receiving more attention in current research on LLMs~\citep{yangwei2024co,yangwei2025candy,yangwei2025elaboration}, ensuring the achievement of robust and reliable AI paradigms~\citep{tian2024crosel,tian2025ddi,pysafedai2026breach}.

\textbf{Table Question Answering.}
Table Question Answering(Table QA) has significant practical applications across various domains, including financial statement analysis~\citep{chen2021finqa} and medical diagnosis~\citep{hasny2025medicaltable}. The field has advanced considerably with the development of high-quality datasets, beginning with the pioneering work of Pasupat et al.~\citep{pasupat2015wtq}, who constructed the WikiTableQuestions (WTQ) dataset using Wikipedia tables. Subsequent research shifted to more complex QA tasks requiring reasoning capabilities, exemplified by datasets such as ToTTo~\citep{parikh2020totto}(focused on answer generation) and OTTQA~\citep{chen2020oTTQA} (emphasizing cross-table reasoning). More recently, FinQA~\citep{chen2021finqa} and AiTQA~\citep{katsis2021aitQA} have explored numerical reasoning in tables, while TableBench~\citep{wu2025tablebench} and Text2Analysis~\citep{he2024text2analysis} introduced multimodal approaches incorporating visual elements. 
There have also been some works in table machine learning that focus on table-type problem solving in open environments~\citep{yu2026thinking,zhou2025ftta}
However, most existing datasets rely on manual annotation, lacking an automated pipeline for scalable data processing, which is common in other application areas~\citep{zhou2025lawgpt,yang2026nesyroute}.

\begin{figure*}[t]
    \centering
    \includegraphics[width=0.85\linewidth]{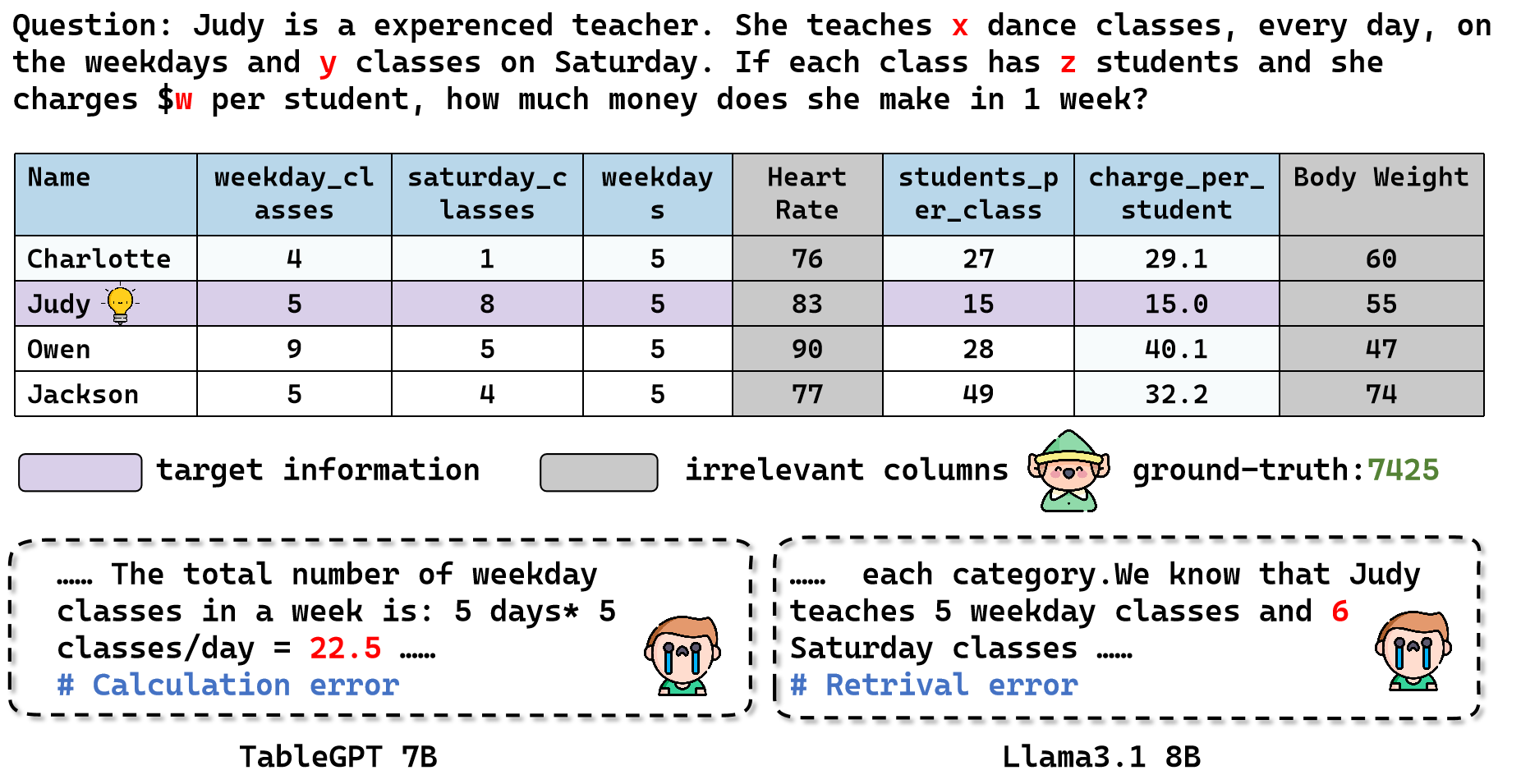}
    \caption{Illustrative cases in \benchmark and corresponding model responses}
    \label{fig:dataset example}
    \vspace{-1.0em}
\end{figure*}

\section{Automated Text to Table}
\label{sec:pipeline}

The \Pipeline pipeline converts math word problems into tabular problems through the following three stages (Fig~\ref{fig: overview pipeline}).

\subsection{Semantic Decoupling}
Firstly, our objective is to semantically decouple the text of the math word problems and extract key elements that can be structurally represented. We decompose math problems using formal language modeling(Such as SMT-Lib~\citep{barrett2010smt,zhi2024neurodata}), structuring problems as:

\vspace{-1em}
\begin{align*}
\textit{Goal} \quad  g := &\text{solve}\ f(v) \\
\textit{Constraints} \quad 
 c := &e_1(v) \bowtie e_2(v), \\
& \bowtie \in \{\geq, \leq, >, <, =, \neq \} \\
\textit{Expressions} \quad 
 e := &h \mid e_1 \oplus e_2, \\
& \oplus \in \{+, -, \times, \div \} \\
\textit{Domains} \quad  \mathcal{D} := &\mathbb{N} \mid \mathbb{N}^+ \mid \mathbb{R}
\end{align*}
\vspace{-1em}


where $v$ is a variable, $c$ is a constraint, $e$ is an expression, $h$ is a constant, and $f$ is the objective function. For a problem $p$, we define the modeling state as $\mathcal{S} = (\mathcal{V}, \mathcal{C})$, where $\mathcal{V}$ and $\mathcal{C}$ denote the variable and constraint sets, respectively.
The LLM constructs $\mathcal{S}$ by extracting candidate components from the problem description, while a formal solver $\Phi$ (e.g., Z3~\citep{Moura2008z3}, CVC5~\citep{barbosa2022cvc5}) verifies satisfiability and consistency, providing feedback for refinement and identifying ill-defined formulations.

\subsection{Table Transformation}

After obtaining the formal modeling state, the next step is to transform the semantically decoupled components into a structured tabular representation. Specifically, we convert the formal state into a table by introducing a \texttt{name} field as the primary key and mapping variables $\mathcal{V}$ and active constraints $\mathcal{C}_a$ to table columns.
Given a problem $p$, the LLM produces a blurred textual description $\hat{p}$ together with a two-row seed table $t_{\text{seed}}$:

\vspace{-1em}
\begin{equation}
        (\hat{p},t_{seed}) = LLM_{tt}(p,\mathcal{V},\mathcal{C}_a)
\end{equation}

To ensure the correctness of entity extraction, we validate the generated table using a formal solver with updated constraints $\mathcal{C}_{\hat{a}}$:
\begin{equation}
    \hat{R}_{valid} =\Phi \left(\mathcal{V}, (\mathcal{C} \setminus \mathcal{C}_a) \cup \mathcal{C}_{\hat{a}}\right)
\end{equation}

\subsection{Table Augmentation} 
Starting from the initial seed table $t_{\text{seed}}$, we construct an augmentation pool $\mathcal{A}$ that applies randomized operations to expand and perturb table structures. The pool consists of four strategies: \textit{Row Augmentation (RowAug)}, \textit{Column Augmentation (ColAug)}, \textit{Order Shuffling (OrdShf)}, and \textit{Information Modification (InfMut)}. Based on user-specified selections, these strategies are iteratively applied to generate diverse table variants.


\begin{equation}\label{Eq:aug}
    t_i = \begin{cases}
    t_{seed} & \text{if }  i = 0 \\
    Aug_j(t_{i-1}) ,\quad Aug_j(\cdot) \in \mathcal{A}   & \text{if } i > 0
    \end{cases}
\end{equation}

The \Pipeline enables automated generation of diverse tabular problem variants without manual annotation. Details on augmentation strategies are in Appendix~\ref{app: augemntation}.

\section{\benchmark Benchmark}

\begin{table*}[t]
\caption{Comparison between \benchmark and existing datasets}
\vspace{-0.5em}
\label{Tab:dataset_comparison}
\resizebox{\linewidth}{!}{
\begin{tabular}{l|cc|cc|ccc|cc}
\toprule
\multirow{3}{*}{Dataset} 
& \multicolumn{2}{c|}{\textbf{Key Statistic}} 
& \multicolumn{2}{c|}{\textbf{Modality}}
& \multicolumn{3}{c|}{\textbf{Content Coverage}} 
& \multicolumn{2}{c}{\textbf{Construction Process}} \\

\cmidrule(lr){2-3} \cmidrule(lr){4-5} \cmidrule(lr){6-8} \cmidrule(lr){9-10}
& Test size 
& \begin{tabular}[c]{@{}c@{}}Table \\ cells\end{tabular}
& Text 
& Image
& \begin{tabular}[c]{@{}c@{}}Information \\ Retrieval\end{tabular}
& \begin{tabular}[c]{@{}c@{}}Math \\ Reasoning\end{tabular}
& Robustness
& \begin{tabular}[c]{@{}c@{}}Automated \\ Verification\end{tabular}
& Symbolizable \\

\midrule
AddSub~\citep{hosseini2014AddSub} 
& 600  & NA 
& \cmark & \xmark
& \xmark & \cmark & \xmark
& \xmark & \xmark \\

SVAMP~\citep{patel2021svamp} 
& 1000 & NA 
& \cmark & \xmark
& \xmark & \cmark & \xmark
& \xmark & \xmark \\

GSM8k~\citep{Cobbe2021gsm8k} 
& 1438 & NA 
& \cmark & \xmark
& \xmark & \cmark & \xmark
& \xmark & \xmark \\

PMC~\citep{tian2025vcsearch} 
& 5374 & NA 
& \cmark & \xmark
& \xmark & \cmark & \cmark
& \cmark & \xmark \\

\midrule
Tabfact~\citep{chen2019tabfact} 
& 1695 & 15.1 
& \cmark & \xmark
& \cmark & \xmark & \xmark
& \xmark & \xmark \\

FinQA~\citep{chen2021finqa} 
& 1147 & 24.5 
& \cmark & \xmark
& \cmark & \cmark & \xmark
& \xmark & \xmark \\

TaT-QA~\citep{zhu2021tatqa} 
& 669  & 37.6 
& \cmark & \xmark
& \cmark & \cmark & \xmark
& \xmark & \xmark \\

TabMWP~\citep{lu2022tabmwp} 
& 7686 & 11.8 
& \cmark & \cmark
& \cmark & \cmark & \xmark
& \xmark & \xmark \\

\midrule
TabularMath
& 6782 & 93.5
& \cmark & \cmark
& \cmark & \cmark & \cmark
& \cmark & \cmark \\

\bottomrule
\end{tabular}
}
\vspace{-1em}
\end{table*}

\label{sec:main dataset}

\begin{table}
\caption{Key statistics in \benchmark}
\label{Tab:Aug in dataset}
\centering
\resizebox{0.5\textwidth}{!}{
\begin{tabular}{l|cccc}
\toprule
Statistic       & Easy & Medium & Hard & Imperfect \\ \hline
Total questions & 797  & 797    & 797  & 1000 \\
Table cells     & 41   & 82     & 162  & 90   \\
Table Rows      & 4.1  & 4.1    & 8.1  & 4.5   \\
Table Columns   & 10   & 20     & 20   & 20   \\
Question Length & 232.2& 232.2  & 232.2& 237   \\ \hline
RowAug          & \cmark & \cmark & \cmark & \cmark \\
OrdShf          & \xmark & \cmark & \cmark & \cmark \\
ColAug          & \xmark & \xmark & \cmark & \xmark \\
InfMut          & \xmark & \xmark & \xmark & \cmark \\
\bottomrule
\end{tabular}
}
\vspace{-1em}
\end{table}


\subsection{Benchmark Details}
To enable standardized evaluation and fair comparison, we construct a predefined benchmark dataset, \benchmark, through the \Pipeline pipeline. We have converted a benchmark, consisting of four subsets: \emph{Easy, Medium, Hard, Imperfect}, covering a total of 3,391 unique tables. For each table, we further augment the data by rendering it into an image format using the matplotlib library in Python and storing the corresponding visual representation. As a result, the final benchmark contains 6,782 samples.
Table~\ref{Tab:Aug in dataset} summarizes the key statistics of each subset as well as the augmentation strategies employed, while additional examples from \benchmark and visual comparisons with GSM8K are provided in Appendix ~\ref{app:data}. We will describe our benchmark subsets separately.



\textbf{\emph{Easy, Medium, Hard} Subset}
For these three subsets, we employ the \Pipeline to systematically reorganize and embed the information originally implicit in the textual descriptions into tabular representations, thereby transforming explicit textual semantics into structured tables. Each subset contains 797 examples (with the remaining samples excluded due to failures in table conversion or consistency validation), and all examples are derived from the same set of seed questions to ensure semantic consistency and comparability across subsets.
The progression from easy to hard reflects an increasing level of table complexity, as measured by the number of rows or columns (induced via \textit{RowAug, ColAug and OrdShf}). This growth in structural complexity substantially increases the difficulty of information localization and retrieval within the tables.

\textbf{\emph{Imperfect} Subset} Real-world tables are not always perfect and may contain errors or omissions. In the \emph{Imperfect} subset, we simulated such a real-world open environment, which falls into two categories: \textbf{missing-type} (removing essential information from the target line of the table) and \textbf{contra-type} (injecting intermediate variables required for the question but designed to conflict with existing information). 
This subset contains 1,000 instances in total, evenly split between the two types (500 per type). These traps are introduced via \emph{Information Modification} during table augmentation.
During evaluation, we inform the model that the table might contain omissions, and observe whether models can accurately identify these traps and abstain from answering, which serves as the core metric. We measure performance by reporting the proportion of ill-defined questions that a model successfully rejects. 


\subsection{Comparison with Existing Benchmarks}

As shown in Table~\ref{Tab:dataset_comparison}, \benchmark differs from existing ones in three key aspects:
(1) Tables in \benchmark are more complex, containing more cells, which makes it harder to retrieve useful information;
(2) Compared to prior mathematical reasoning datasets, our benchmark jointly evaluates reasoning and retrieval abilities, and includes an imperfect subset with incomplete information. Unlike existing tabular QA datasets, we emphasize mathematical reasoning and assess models’ ability to detect traps (e.g., flawed or contradictory conditions), promoting robust and safe reasoning.
(3) In terms of construction, unlike prior work that relies heavily on manual annotation, we adopt a neuro-symbolic approach  \Pipeline, which rewrites textual problems into tabular form. This allows us to generate multiple table variants for the same seed problem, achieving efficient and controllable data creation.


\begin{table}[t]
\caption{Main results on \benchmark benchmark}
\label{tab:main_results_left}
\centering
\small
\setlength{\tabcolsep}{2pt}
\begin{tabular}{lccccc}
\toprule
Dataset & GSM8k & Easy & Medium & Hard & Avg \\
\midrule
\multicolumn{6}{l}{\textbf{Open-source General Models}} \\
\midrule
Qwen3 14B        & \textbf{94.54} & 77.87 & 70.21 & 61.59 & \textbf{69.89} \\
Qwen3 8B         & \textbf{93.30} & 73.18 & 55.63 & 47.30 & \textbf{58.70} \\
Qwen3 4B         & \textbf{91.79} & 71.57 & 52.68 & 41.73 & \textbf{55.32} \\
Qwen3 1.7B         & \textbf{81.25} & 50.76 & 30.62 & 18.84 & \textbf{33.40} \\
Qwen2.5 14B      & \textbf{93.40} & 79.21 & 64.10 & 49.09 & \textbf{64.13} \\
Qwen2.5 14B coder& \textbf{90.68} & 72.63 & 57.74 & 45.61 & \textbf{58.66} \\
Qwen2.5 7B       & \textbf{82.86} & 53.92 & 34.45 & 20.64 & \textbf{36.33} \\
Qwen2.5 7B coder & \textbf{84.71} & 64.78 & 42.13 & 23.52 & \textbf{43.47} \\
Qwen2.5 3B       & \textbf{80.28} & 39.74 & 23.96 & 15.68 & \textbf{26.46} \\
LLaMA3.1 8B      & \textbf{83.69} & 48.61 & 33.37 & 32.15 & \textbf{38.04} \\
LLaMA3 8B      & \textbf{55.34} & 36.30 & 21.12 & 20.68 & \textbf{26.03} \\
\midrule
\multicolumn{6}{l}{\textbf{Open-source Math Models}} \\
\midrule
Qwen-Math 7B     & \textbf{95.45} & 53.69 & 30.37 & 14.59 & \textbf{32.88} \\
DeepSeek-Math 7B & \textbf{80.13} & 12.81 & 6.60  & 2.04  & \textbf{7.15}  \\
\midrule
\multicolumn{6}{l}{\textbf{Open-source Tabular Models}} \\
\midrule
TableGPT 7B      & \textbf{24.33} & 30.60 & 16.44 & 17.64 & \textbf{21.56} \\
StructLM 7B      & \textbf{32.97} & 14.78 & 8.28  & 4.44  & \textbf{9.17}  \\
\midrule
\multicolumn{6}{l}{\textbf{Closed-source APIs}} \\
\midrule
DeepSeek v3      & \textbf{96.36} & 88.63 & 87.63 & 85.83 & \textbf{87.37} \\
GLM-4-Plus       & \textbf{95.07} & 84.52 & 81.03 & 78.27 & \textbf{81.27} \\
GPT-4            & \textbf{94.46} & 85.54 & 78.42 & 75.23 & \textbf{79.73} \\
\bottomrule
\end{tabular}
\end{table}

\section{Experiments and Results}
In this section, we conduct a series of experiments on 18 models spanning four categories to analyze their performance on mathematical reasoning over structured data systematically. We primarily investigate the following three research questions:
(1)How does table complexity affect mathematical reasoning?
(2)How does table quality affect mathematical reasoning?
(3)How does table representation affect mathematical reasoning?

\subsection{RQ1. How does table complexity affect mathematical reasoning?}

\textbf{Performance drops when moving from pure text to tabular modalities, and increases in table complexity further exacerbate this degradation.} 
As shown in Table~\ref{tab:main_results_left}, when transitioning from the original GSM8K dataset to our constructed \benchmark, all models exhibit consistent and significant performance degradation. Moreover, this degradation becomes more pronounced as table complexity increases. Smaller models, whose reasoning capabilities are inherently limited, suffer more severe performance drops, while domain-specific models (including math-specialized and table-specialized models) also struggle to generalize under this setting.
We further conduct manual analysis on a subset of model outputs and identify four primary error types: retrieval omission (failing to recognize the need for retrieval), retrieval mismatch (retrieving incorrect table values), expression errors (incorrect formulation of the target equation), and numerical calculation errors (correct formulas but incorrect computations). Among these, retrieval mismatch is the dominant source of errors. Detailed case studies are provided in Appendix~\ref{app:caseanalysis}.

\begin{figure}[t]
    \centering
    \includegraphics[width=\linewidth]{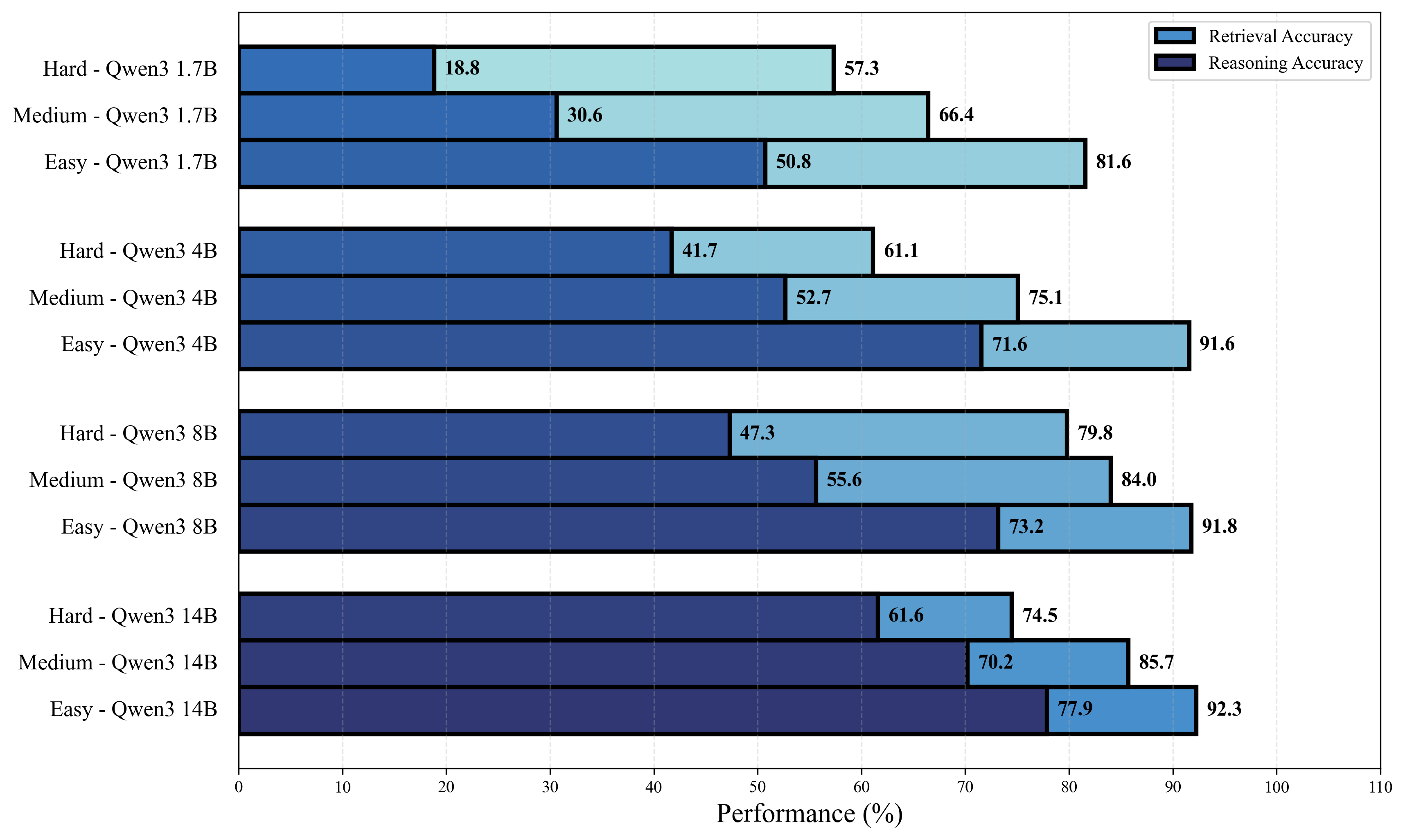}
    \vspace{-2em}
    \caption{Performance comparison of table reasoning and single-step table retrieval}
    \label{fig: findings 2}
\vspace{-1.5em}
\end{figure}

\textbf{\textit{Is the degradation in model inference performance due to the inability to find information? }Pure retrieval is substantially easier than joint reasoning and retrieval.}  
Pure retrieval refers to replacing end-to-end reasoning with a simplified single-step retrieval task. Specifically, given a question such as \texttt{"Janet's ducks lay x eggs per day ..."}, we directly ask \texttt{"How many eggs do Janet's ducks lay per day?"}, which corresponds to the \texttt{"eggs-per-day"} key in the table.As shown in Fig.~\ref{fig: findings 2}, model performance on full reasoning tasks is significantly worse than on pure retrieval tasks. This indicates that when the retrieval target is explicitly specified, models can reliably perform retrieval. However, once retrieval is embedded within multi-step reasoning, performance drops by an average of 20\%. This reveals a substantial gap between retrieval and reasoning over tables. 

\begin{figure}[t]
    \centering
    \includegraphics[width=\linewidth]{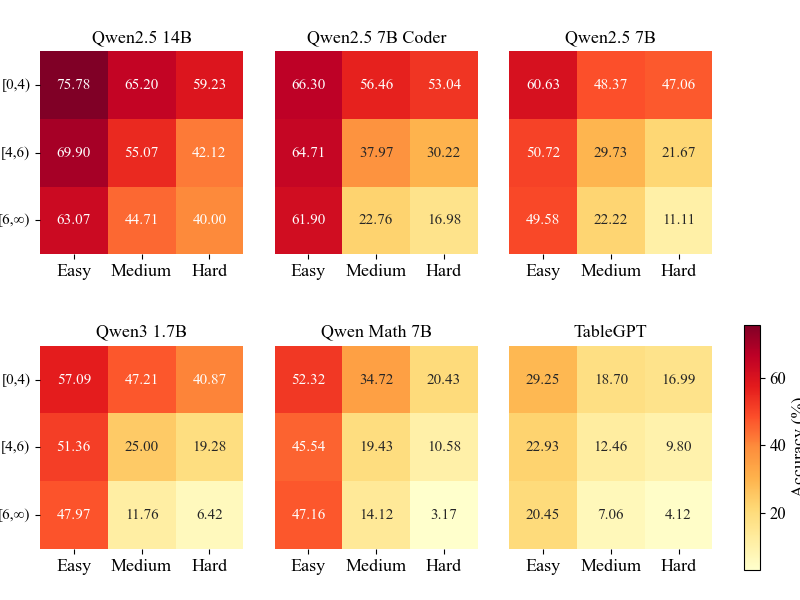}
    \vspace{-2em}
    \caption{Accuracy heatmap of table complexity and reasoning difficulty in tabular math reasoning}
    \label{fig: heatmap}
    \vspace{-1.5em}
\end{figure}

Moreover, we categorize all test samples in \benchmark for each model along two dimensions: retrieval difficulty (i.e., table complexity, divided into Easy, Medium, and Hard subsets) and reasoning difficulty (measured by the number of variables). This results in nine categories, whose accuracies are visualized as a heatmap in Fig.~\ref{fig: heatmap}. The results indicate that model performance is strongly correlated with both factors, and the impacts of retrieval difficulty and reasoning difficulty are nearly comparable. 


\subsection{RQ2. How does table quality affect mathematical reasoning?}

\begin{table}[t]
\caption{Performance on Imperfect subset}
\label{tab:main_results_robust}
\centering
\scriptsize
\setlength{\tabcolsep}{3pt}
\small
\begin{tabular}{lccc}
\toprule
Dataset & Missing & Contradictory & Average \\
\midrule
\multicolumn{4}{l}{\textbf{Open-source General Models}} \\
\midrule
Qwen3 14B        & 67.03 & 28.21 & \textbf{47.62} \\
Qwen3 8B         & 68.57 & 34.54 & \textbf{51.56} \\
Qwen3 4B         & 69.75 & 32.12 & \textbf{50.94} \\
Qwen3 1.7B       & 37.33 & 14.40 & \textbf{29.56} \\
Qwen2.5 14B      & 20.00 & 6.80  & \textbf{13.40} \\
Qwen2.5 14B Coder& 51.60 & 23.60 & \textbf{37.60} \\
Qwen2.5 7B       & 34.80 & 16.00 & \textbf{25.40} \\
Qwen2.5 7B Coder & 34.00 & 20.80 & \textbf{27.40} \\
Qwen2.5 3B       & 79.20 & 69.60 & \textbf{74.40} \\
LLaMA3.1 8B      & 9.60  & 10.40 & \textbf{10.00} \\
LLaMA3 8B        & 35.20 & 19.20 & \textbf{21.90} \\
\midrule
\multicolumn{4}{l}{\textbf{Open-source Math Models}}\\
\midrule
Qwen-Math 7B     & 48.93 & 20.80 & \textbf{34.87} \\
DeepSeek-Math 7B& 53.60 & 60.40 & \textbf{57.00} \\
\midrule
\multicolumn{4}{l}{\textbf{Open-source Tabular Models}}\\
\midrule
TableGPT 7B      & 46.40 & 23.20 & \textbf{34.80} \\
StructLM 7B      & 0.00     & 0.00     & \textbf{0.00} \\
\midrule
\multicolumn{4}{l}{\textbf{Closed-source APIs}} \\
\midrule
DeepSeek v3      & 82.80 & 68.00 & \textbf{75.40} \\
GLM-4-plus       & 65.60 & 27.60 & \textbf{46.60} \\
GPT-4            & 80.20 & 21.11 & \textbf{50.66} \\
\bottomrule
\end{tabular}
\end{table}

The quality of tables is not always guaranteed to be intact; information may be missing or contradictory. This research question aims to investigate whether, under such circumstances, models can timely detect these issues and inform users that the question is ill-posed or problematic, rather than providing a misleading answer. 

\textbf{Flawed table quality could pose significant risks for current LLMs.}
As shown in~\ref{tab:main_results_robust}, when tables contain missing or contradictory information, most LLMs fail to properly identify these flaws and instead produce misleading answers. The high Missing and Contradictory scores indicate that models frequently generate responses even when critical information is absent or conflicting, rather than alerting users to the problematic nature of the questions. This behavior is particularly dangerous in real-world applications where users may rely on these answers without recognizing the underlying data quality issues. Contradictory cases prove more challenging than missing-information cases, suggesting that models struggle more when information conflicts rather than when it is simply absent. Even advanced proprietary APIs exhibit significant vulnerabilities, with average error rates exceeding 50\% in some cases. Notably, the seemingly higher robustness of some smaller models mainly stems from their coarse-grained refusal behavior, as they tend to reject most queries rather than genuinely reasoning about flawed tables.

\begin{figure}[t]
    \centering
    \includegraphics[width=0.85\linewidth]{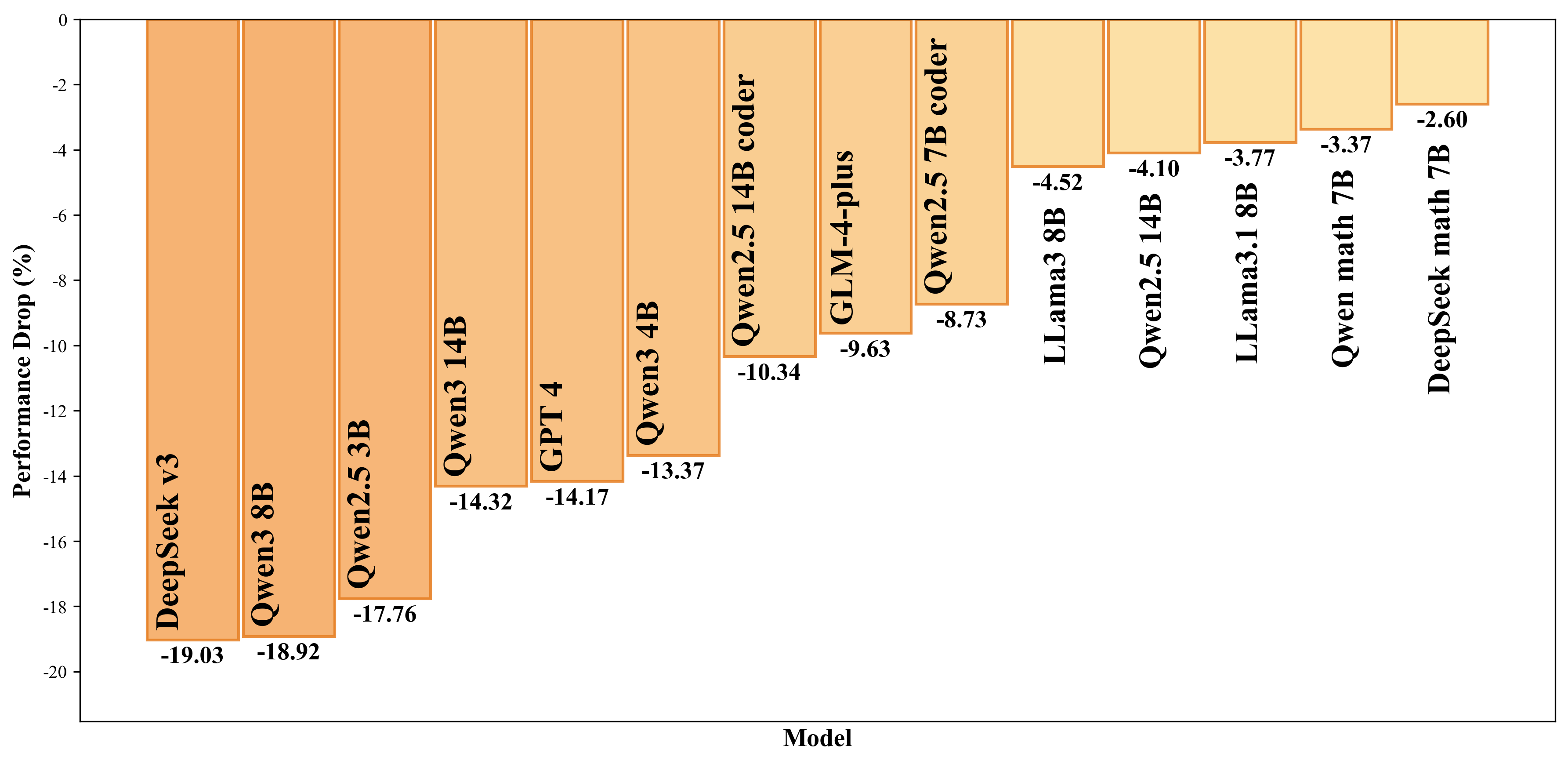}
    \caption{Performance degradation on well-defined table problems caused by table-checking instructions}
    \label{fig: trade-off}
    \vspace{-1em}
\end{figure}

\textbf{\textit{What impact does informing the model of potential issues with table formatting have on solving normal well-defined problems?} All models experienced varying degrees of performance degradation, demonstrating the trade-off between solvency and discriminative ability.} 
When models are explicitly informed that inputs may contain trap problems, they must first assess problem solvability before solution planning. This discriminative requirement forces models to allocate resources to verification, altering the reasoning process. As shown in Table~\ref{fig: trade-off}, this mixed setting leads to performance degradation on well-defined problems across all model types, including advanced proprietary APIs. The trade-off between solvency and discriminative ability indicates that current models struggle to simultaneously maintain high accuracy on solvable problems while effectively detecting problematic inputs.



\subsection{RQ3. How does table representation affect mathematical reasoning?}
Given that real-world scenarios frequently involve diverse forms of image-based tables, a natural and important question arises as to how reasoning over image-based tables compares to reasoning over text-based tables across different models
\begin{table}[t]
    \caption{Performance comparison with and without visual table representations on \benchmark}
    \label{tab:modality_results}
    \centering
    \small
    \setlength{\tabcolsep}{2pt}
    \begin{tabular}{lccccc}
    \toprule
    \multirow{2}{*}{Model} & \multicolumn{4}{c}{with image} & w/o \\
    \cmidrule(lr){2-5} \cmidrule(lr){6-6}
     & Easy & Medium & Hard & Avg & Avg \\
    \midrule
    Qwen2.5-VL-7B   & 57.26 & 49.36 & 42.70 & \textbf{49.44} & 45.51 \\
    Qwen2.5-VL-3B   & 29.65 & 23.36 & 18.26 & 23.76 & \textbf{26.47} \\
    Qwen3-VL-8B     & 66.25 & 64.14 & 59.83 & 63.41 & \textbf{66.40} \\
    Qwen3-VL-4B     & 66.17 & 60.22 & 51.22 & 59.20 & \textbf{60.09} \\
    InternVL3-8B    & 15.69 & 14.20 & 10.02 & \textbf{13.30} & 10.82 \\
    InternVL3-14B   & 52.60 & 34.03 & 20.13 & 35.59 & \textbf{38.66} \\
    \bottomrule
    \end{tabular}
    \vspace{-1em}
    \end{table}

\textbf{Image-based and text-based tables exhibit similar performance trends, but image-based tables are more challenging to reason over.}
As shown in Table~\ref{tab:modality_results}, across different models and difficulty levels, image-based tables follow performance trends largely consistent with those of text-based tables, with accuracy decreasing as problem difficulty increases. However, even for multimodal models, accuracy on text-based tables is comparable to or higher than that on image-based tables, with gains of 0.89--3.07 percentage points. These results suggest that image-based tables do not necessarily provide more usable information for mathematical reasoning, likely due to OCR noise and layout ambiguity, whereas text-based representations offer more explicit structure and numerical precision.



\begin{figure}[t]
    \centering
    \includegraphics[width=0.85\linewidth]{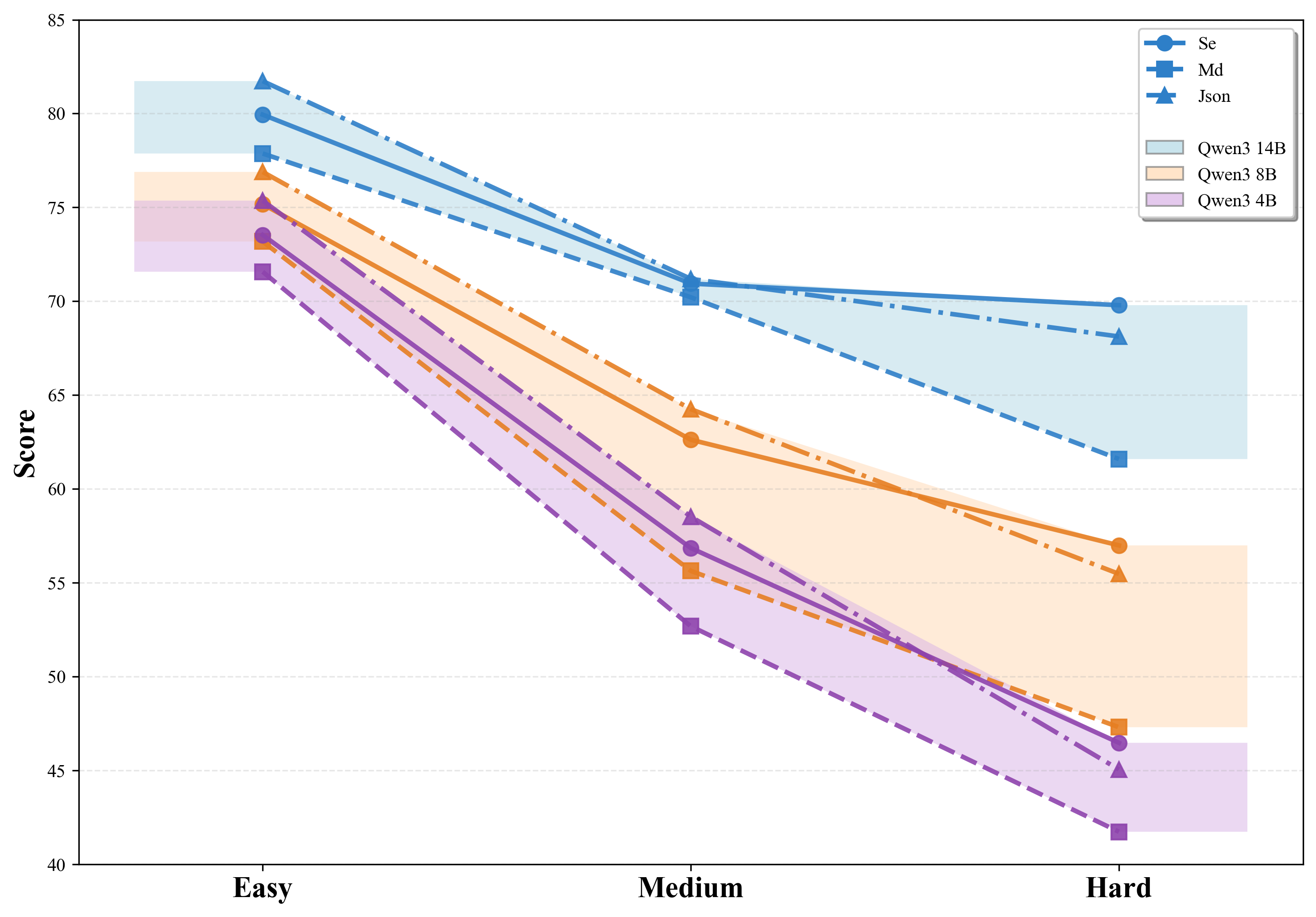}
    \caption{Performance comparison across different table formats (Se, Md, Json) and difficulty levels}
    \label{fig: se_md_mean}
\end{figure}

\textbf{\textit{For text-based tables, does the choice of table format affect model performance across different difficulty levels?} JSON and Serialized formats consistently outperform Markdown, with the gap widening as table complexity increases.} 
We evaluate three widely used text-based table formats representations: serialized, Markdown, and JSON, and conduct a detailed comparison of their performance (Explanation in Appendix).
As shown in Fig.~\ref{fig: se_md_mean}, JSON and serialized representations achieve comparable performance and consistently outperform Markdown across all difficulty levels(Full results in table~\ref{Table: main results appendix}). Moreover, the performance gap becomes more pronounced as problem complexity increases. We attribute this trend to the explicit key–value structure in JSON and serialized formats, where each row is associated with clear keys, facilitating more reliable retrieval and reasoning, whereas Markdown introduces additional parsing ambiguity that degrades performance on complex tables.

\subsection{Further Discussion}

\begin{figure}[t]
    \centering
    \includegraphics[width=\linewidth]{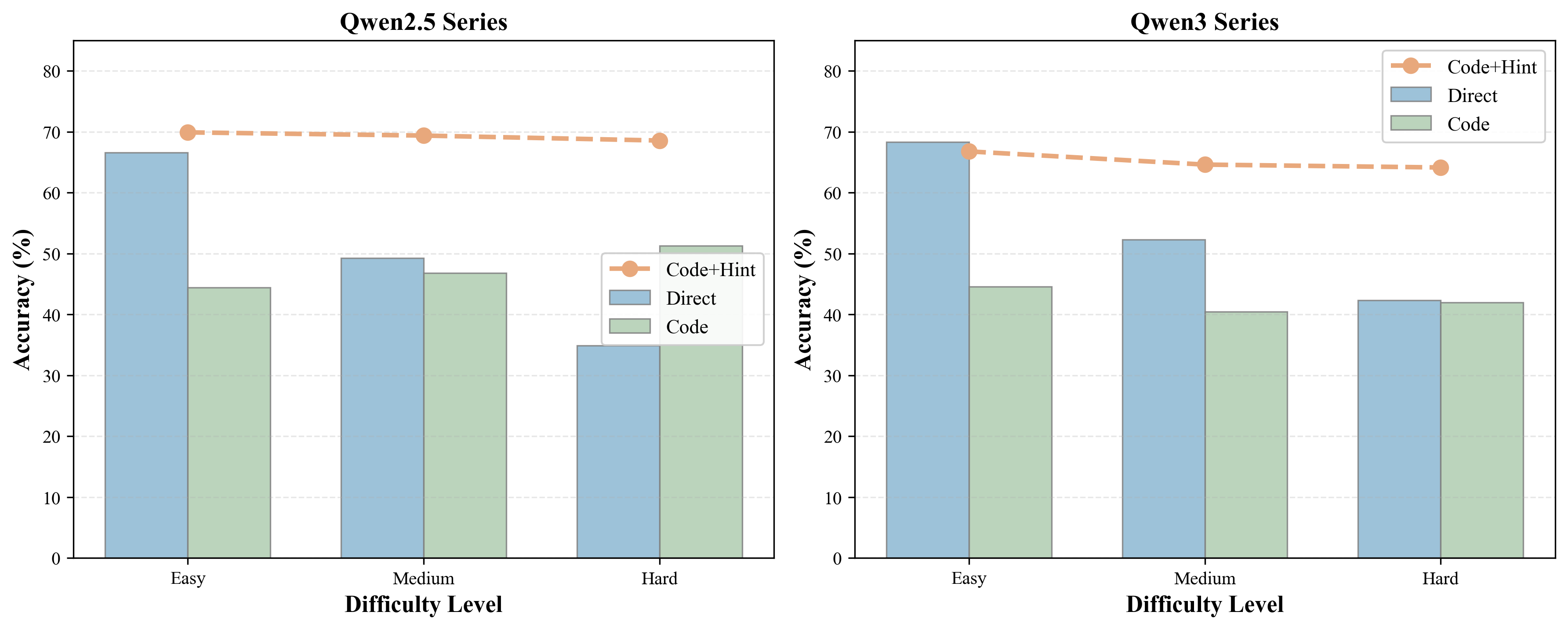}
    \caption{Performance comparison between code-based reasoning and direct reasoning.}
    \label{fig: Code comparison}
    \vspace{-1em}
\end{figure}



Beyond above questions, we further explore several complementary aspects, including code-based solving (Fig.~\ref{fig: Code comparison}), fine-tuning performance (Table~\ref{Tab:train:ID}), row/column ablations (Fig.~\ref{fig: colaug/rowaug ablation}), and trap-type ablations (Table~\ref{tab: trap type}). Due to space limitations, some analyses are deferred to the appendix.

\textbf{\textit{Can using code enhance the model's inference ability?} 
Models often struggle to accurately infer what information should be retrieved or executed during the reasoning process.} This limitation constrains the potential benefits of code-based reasoning. We conduct systematic experiments on the Qwen2.5 and Qwen3 series of models, and visualize in  Fig~\ref{fig: Code comparison} (Full Results in Table~\ref{tab:code comparation}). The results show that incorporating code leads to more stable performance overall; however, in many cases, it does not outperform direct natural language reasoning.
In contrast, when the prompt explicitly specifies which pieces of information should be retrieved and used for reasoning, model performance improves substantially. These findings suggest that the performance gains stem not from the mere inclusion of code, but from providing the model with a clear and actionable information retrieval and reasoning trajectory.


\begin{table}[t]
    \centering
    \caption{Performance changes on \benchmark of the basic model after \Pipeline training (In-Domain)}
    \label{Tab:train:ID}
    \resizebox{\linewidth}{!}{
    \begin{tabular}{lccc}
    \toprule
    Settings    & Easy (Md) & Medium (Md) & Hard (Md) \\
    \midrule
    Qwen2.5-3B-Instrct            & 39.74 & 23.96 & 15.68 \\
    Qwen2.5-3B-Instrct + finetune & 53.58 & 41.78 & 36.51 \\
    \rowcolor{black!15}
    $\Delta$                      & \textcolor{red}{$\uparrow$ 13.84} & \textcolor{red}{$\uparrow$ 17.82} & \textcolor{red}{$\uparrow$ 20.83} \\
    \midrule
    Qwen2.5-7B-Instrct            & 53.92 & 34.45 & 20.64 \\
    Qwen2.5-7B-Instrct + finetune & 62.10 & 51.32 & 36.39 \\
    \rowcolor{black!15}
    $\Delta$                      & \textcolor{red}{$\uparrow$ 8.18} & \textcolor{red}{$\uparrow$ 16.87} & \textcolor{red}{$\uparrow$ 15.75} \\
    \bottomrule
    \end{tabular}
    }
    \vspace{-1em}
    \end{table}

\textbf{\textit{How about using \Pipeline-generated data for training?} A flexible data generation framework can provide substantial support for improving tabular mathematical reasoning.}
To improve tabular mathematical reasoning, we leverage \Pipeline to generate targeted training data. Starting from the GSM8K training set, we construct approximately 6K augmented samples and fine-tune Qwen2.5-7B and Qwen2.5-3B. As shown in Table~\ref{Tab:train:ID}, fine-tuning yields an average improvement of about 15\% on \benchmark. These gains further transfer to other tabular reasoning benchmarks (TAT-QA, FinQA, and TabMWP), with an average improvement of 4\%(Table~\ref{tab:train OOD}), particularly on complex tables, demonstrating the effectiveness of controllable data generation.


\section{Conclusion}



In this work, we introduce \benchmark, a comprehensive benchmark for systematically studying tabular mathematical reasoning in LLMs. It comprises three subsets of increasing complexity and an imperfect subset to evaluate both performance and robustness. We further propose a neural–symbolic framework, \Pipeline, which controllably transforms word problems into scalable and validated tabular reasoning tasks. Extensive experiments on 18 models reveal key challenges in tabular reasoning from the perspectives of table complexity, table quality, and table representation, and offer insights for future research.

\section*{Limitations.} 
First, the seed dataset GSM8k used to construct \benchmark is relatively homogeneous, as all derived instances originate from a limited set of mathematical word problems. Although our transformation pipeline generates diverse tabular structures and difficulty levels, the underlying semantic diversity of problem contexts remains constrained, which may limit the generality of our findings.
Second, while \Pipeline effectively scales task difficulty by increasing retrieval complexity (e.g., table complexity, structure, and information distribution), it places less emphasis on systematically increasing intrinsic reasoning complexity, such as deeper multi-step symbolic dependencies. As a result, the benchmark primarily stresses retrieval and alignment capabilities rather than fully isolating limitations in complex mathematical reasoning.

\section*{Acknowledgements}
This research was supported by the Jiangsu Science Foundation (BK20232003, BK20243012, BG2024036), Natural Science Foundation of China (624B2068, 62576162), and the Fundamental Research Funds for the Central Universities (022114380023).

\bibliography{custom}

\appendix
\section{Appendix}

\subsection{Use of LLMs}
This work investigates the boundaries of large language models (LLMs) in tabular mathematical reasoning. In this process, LLMs serve a dual role. On the one hand, they act as both the subject of study and a tool for dataset construction and experimental evaluation, supporting data generation (Sec.~3) and benchmarking representative models (Sec.~4). On the other hand, LLMs are additionally employed to polish the writing and enhance the clarity of English expression. 

All key research ideas, theoretical analysis, experimental design, and writing of the main body of the paper were independently completed by the authors. We did not use the large language model to generate the scientific content of the manuscript, nor did we contribute to the formulation of the research hypotheses or the interpretation of the findings. The authors bear full responsibility for the accuracy, originality, and completeness of all content in the paper.

\subsection{Details of TableGsm8k Dataset}
\label{app:data}
To evaluate the reasoning ability of the models on structured data, we construct the \benchmark benchmark including four subsets: \emph{Easy, Medium, Hard,} and \emph{Imperfect}.The \emph{Imperfect} subset contains 50\% solvable problems (corresponding to medium difficulty) and 50\% unsolvable problems (25\% with contradictory conditions + 25\% with missing information).\\\\
Taking the following problem as an example (Example~\ref{subsets example 1}), we use the four categories mentioned in 3.1.3 to generate five tables(Table~\ref{tab:table 7}~\ref{tab:table 8}~\ref{tab:table 9}~\ref{tab:table 10}~\ref{tab:table 11}) based on the seed row and corresponding generalized problem, so that the table and generalized problem form a question pair as our dataset.
\begin{itemize}
    \item \textbf{Easy:} Apply \textcolor{blue}{RowAug} (10 times) and \textcolor{blue}{Shuffling} to the seed row

    \item \textbf{Medium:} Apply \textcolor{blue}{RowAug} (20 times), \textcolor{blue}{Shuffling} to the seed row
    
    \item \textbf{Hard:} Apply \textcolor{blue}{RowAug} (20 times), \textcolor{blue}{Shuffling}, and \textcolor{blue}{ColAug} (4 columns) (adding irrelevant information marked with gray)
    
    \item \textbf{Imperfect:} Apply \textcolor{blue}{RowAug} (20 times), \textcolor{blue}{Shuffling}, and \textcolor{blue}{InfMod}(two situations) to the seed row
    \begin{itemize}
         \item \textbf{Contra:} Apply \textcolor{blue}{Contradictory Condition Modification}, adding row \textit{"eggs\_for\_sale"}, which is an implicit variable that can be obtained from the formula "\textit{eggs\_for\_sale = eggs\_per\_day - eggs\_eaten - eggs\_for\_muffins}". Modify this implicit variable (original value is marked with blue) to create conflicts with existing constraints, making the problem unsolvable.
    
        \item \textbf{Missing:} Apply \textcolor{blue}{Missing Condition Modification}, removing a key data (marked with yellow)(set as null) from seed row.
    \end{itemize}
\end{itemize}

\begin{figure*}[ht]
    \begin{definedbox}[label=subsets example 1]{Example 1}
    \textbf{Original Problem:} Janet’s ducks lay \textcolor{red}{16} eggs per day. She eats \textcolor{red}{three} for breakfast every morning and bakes muffins for her friends every day with \textcolor{red}{four}. She sells the remainder at the farmers' market daily for \$\textcolor{red}{2} per fresh duck egg. How much in dollars does she make every day at the farmers' market?
    
    \vspace{0.5em}\hdashrule{\linewidth}{0.5pt}{2pt}\vspace{0.5em}
    
    \textbf{Generalized Problem:} Janet’s ducks lay \textcolor{red}{x} eggs per day. She eats \textcolor{red}{y} for breakfast every morning and bakes muffins for her friends every day with \textcolor{red}{z}. She sells the remainder at the farmers' market daily for \$\textcolor{red}{w} per fresh duck egg. How much in dollars does she make every day at the farmers' market?
    
    \vspace{0.5em}\hdashrule{\linewidth}{0.5pt}{2pt}\vspace{0.5em}
    
    \textbf{Seed Row:} (red word in table)
    \begin{center}
    \begin{tabular}{lrrrr}
    \hline
    \textbf{Name} & \textbf{Eggs per Day} & \textbf{Eggs Eaten} & \textbf{Eggs for Muffins} & \textbf{Price per Egg} \\
    \hline
    Janet & 16 & 3 & 4 & 2 \\
    \hline
    \end{tabular}
    \end{center}
    \end{definedbox}
\end{figure*}

\newpage

\begin{table*}[ht]
\centering
\caption{Easy Table}
\label{tab:table 7}
\begin{tabular}{lrrrr}
\toprule
Name & Eggs per Day & Eggs Eaten & Eggs for Muffins & Price per Egg \\
\midrule
Sebastian & 72 & 8 & 16 & 7 \\
Sofia & 73 & 1 & 17 & 7 \\
Elijah & 5 & 4 & 14 & 10 \\
Mia & 73 & 9 & 19 & 7 \\
Ava & 46 & 0 & 7 & 5 \\
Samuel & 3 & 8 & 6 & 7 \\
Logan & 47 & 0 & 9 & 7 \\
Henry & 95 & 9 & 9 & 6 \\
\textcolor{red}{Janet} & \textcolor{red}{16} & \textcolor{red}{3} & \textcolor{red}{4} & \textcolor{red}{2} \\
Ella & 65 & 8 & 15 & 4 \\
Elizabeth & 54 & 0 & 2 & 4 \\
\bottomrule
\end{tabular}
\end{table*}

\begin{table*}[ht]
\centering
\caption{Medium Table}
\label{tab:table 8}
\begin{tabular}{rrlrr}
\toprule
Price/Egg & For Muffins & Name & Eaten & Eggs/Day \\
\midrule
1 & 18 & Jacob & 1 & 17 \\
8 & 7 & Sebastian & 0 & 20 \\
9 & 12 & Lillian & 3 & 59 \\
6 & 14 & Aiden & 1 & 80 \\
7 & 20 & Joseph & 2 & 11 \\
7 & 4 & James & 9 & 20 \\
9 & 16 & Grace & 4 & 36 \\
10 & 17 & Mia & 6 & 90 \\
8 & 8 & Oliver & 3 & 43 \\
6 & 13 & Charlotte & 0 & 18 \\
9 & 3 & Mia & 3 & 79 \\
10 & 15 & Mason & 1 & 34 \\
1 & 1 & Jacob & 9 & 85 \\
4 & 8 & Lucas & 8 & 95 \\
3 & 13 & Liam & 8 & 56 \\
10 & 5 & James & 8 & 84 \\
1 & 0 & Oliver & 5 & 47 \\
6 & 19 & Eleanor & 5 & 40 \\
3 & 10 & Victoria & 9 & 68 \\
5 & 5 & Samuel & 5 & 16 \\
\textcolor{red}{2} & \textcolor{red}{4} & \textcolor{red}{Janet} & \textcolor{red}{3} & \textcolor{red}{16} \\
\bottomrule
\end{tabular}
\end{table*}

\newpage

\begin{table*}[ht]
\centering
\caption{Hard Table}
\label{tab:table 9}
\begin{tabular}{>{\columncolor{gray!30}}r>{\columncolor{gray!30}}rrrlrr>{\columncolor{gray!30}}r>{\columncolor{gray!30}}r}
\toprule
Age & Heart Rate & Eggs/Day & Price/Egg & Name & Eaten & For Muffins & Body Temp & Sleep Hours \\
\midrule
23 & 81 & 75 & 1 & Emma & 3 & 1 & 38 & 4 \\
46 & 79 & 81 & 5 & Chloe & 7 & 0 & 38 & 9 \\
43 & 88 & 10 & 3 & Emma & 2 & 14 & 39 & 6 \\
42 & 73 & 41 & 9 & Madison & 3 & 11 & 40 & 10 \\
51 & 87 & 98 & 3 & Eleanor & 0 & 4 & 40 & 9 \\
50 & 97 & 94 & 6 & Olivia & 1 & 7 & 37 & 10 \\
63 & 67 & 93 & 3 & Lily & 5 & 16 & 40 & 8 \\
38 & 70 & 51 & 5 & David & 5 & 11 & 39 & 6 \\
70 & 87 & 19 & 10 & Isabella & 3 & 17 & 40 & 7 \\
64 & 99 & 11 & 1 & Avery & 8 & 9 & 38 & 10 \\
72 & 67 & 81 & 7 & Emily & 1 & 20 & 38 & 4 \\
57 & 69 & 38 & 6 & Ella & 7 & 16 & 36 & 4 \\
25 & 62 & 94 & 10 & John & 3 & 11 & 39 & 5 \\
71 & 91 & 29 & 5 & Camila & 6 & 7 & 38 & 4 \\
42 & 73 & 62 & 9 & Layla & 7 & 17 & 36 & 8 \\
62 & 96 & 32 & 7 & Harper & 2 & 19 & 38 & 6 \\
36 & 78 & 77 & 8 & Olivia & 6 & 3 & 39 & 9 \\
48 & 85 & 7 & 7 & Aiden & 8 & 10 & 38 & 8 \\
60 & 82 & 20 & 5 & Joseph & 9 & 19 & 38 & 6 \\
30 & 94 & 77 & 7 & Logan & 2 & 18 & 40 & 7 \\
\textcolor{red}{25} & \textcolor{red}{72} & \textcolor{red}{16} & \textcolor{red}{2} & \textcolor{red}{Janet} & \textcolor{red}{3} & \textcolor{red}{4} & \textcolor{red}{39} & \textcolor{red}{5} \\
\bottomrule
\end{tabular}
\end{table*}

\begin{table*}[ht]
\centering
\caption{Table with Contradictory Conditions}
\label{tab:table 10}
\begin{tabular}{rrrlrr}
\toprule
eggs\_per\_day & \makecell{eggs\_for\_sale\\ \textcolor{blue}{(real\_eggs\_for\_sale)}} & eggs\_eaten & name & eggs\_for\_muffins & price\_per\_egg \\
\midrule
65 & 10 \textcolor{blue}{(38)} & 7 & Noah & 20 & 4 \\
87 & 9 \textcolor{blue}{(65)} & 5 & Wyatt & 17 & 1 \\
95 & 13 \textcolor{blue}{(83)} & 0 & Jayden & 12 & 1 \\
47 & 13 \textcolor{blue}{(27)} & 8 & Lucas & 12 & 4 \\
34 & 15 \textcolor{blue}{(18)} & 9 & Ethan & 7 & 7 \\
72 & 13 \textcolor{blue}{(53)} & 7 & Liam & 12 & 2 \\
79 & 8 \textcolor{blue}{(53)} & 10 & Sofia & 16 & 4 \\
12 & 7 \textcolor{blue}{(-1)} & 5 & Lily & 8 & 9 \\
58 & 13 \textcolor{blue}{(45)} & 5 & Sophia & 8 & 1 \\
31 & 12 \textcolor{blue}{(28)} & 0 & Jayden & 3 & 5 \\
90 & 12 \textcolor{blue}{(78)} & 10 & Ava & 2 & 3 \\
86 & 16 \textcolor{blue}{(58)} & 8 & Sophia & 20 & 10 \\
45 & 14 \textcolor{blue}{(42)} & 1 & Amelia & 2 & 8 \\
44 & 16 \textcolor{blue}{(37)} & 7 & Victoria & 0 & 10 \\
84 & 10 \textcolor{blue}{(64)} & 7 & Mason & 13 & 9 \\
\textcolor{red}{16} & \textcolor{red}{12} \textcolor{blue}{(7)} & \textcolor{red}{3} & \textcolor{red}{Janet} & \textcolor{red}{4} & \textcolor{red}{2} \\
74 & 7 \textcolor{blue}{(60)} & 7 & Oliver & 7 & 10 \\
43 & 15 \textcolor{blue}{(31)} & 3 & Aiden & 9 & 8 \\
82 & 16 \textcolor{blue}{(70)} & 7 & Michael & 5 & 5 \\
57 & 16 \textcolor{blue}{(45)} & 0 & Riley & 12 & 7 \\
52 & 12 \textcolor{blue}{(31)} & 5 & Henry & 16 & 8 \\
\bottomrule
\end{tabular}
\end{table*}

\newpage

\begin{table*}[ht]
\centering
\caption{Table with Missing Information}
\label{tab:table 11}
\begin{tabular}{rrlrr}
\toprule
eggs\_per\_day & eggs\_eaten & name & price\_per\_egg & eggs\_for\_muffins \\
\midrule
66 & 2 & Riley & 8 & 10 \\
97 & 2 & Hannah & 3 & 0 \\
70 & 3 & Olivia & 3 & 20 \\
51 & 8 & Charlotte & 6 & 9 \\
79 & 0 & Elizabeth & 2 & 13 \\
\textcolor{red}{16} & \cellcolor{yellow!50}\textcolor{red}{null} & \textcolor{red}{Janet} & \textcolor{red}{2} & \textcolor{red}{4} \\
14 & 10 & Ava & 8 & 3 \\
48 & 4 & Ethan & 3 & 14 \\
73 & 7 & Olivia & 3 & 20 \\
32 & 0 & Chloe & 3 & 14 \\
41 & 8 & James & 3 & 0 \\
1 & 1 & Benjamin & 3 & 4 \\
8 & 0 & Sophia & 3 & 13 \\
20 & 6 & Victoria & 8 & 14 \\
93 & 10 & John & 9 & 8 \\
62 & 0 & Penelope & 10 & 10 \\
21 & 2 & Harper & 6 & 5 \\
17 & 1 & Oliver & 10 & 10 \\
60 & 3 & John & 4 & 4 \\
14 & 0 & David & 9 & 3 \\
76 & 0 & Jayden & 1 & 7 \\
\bottomrule
\end{tabular}
\end{table*}

\subsubsection{Detailed explanation of augmentations}
\label{app: augemntation}
The detailed explanation of augmentations used in $\mathcal{A}$ are as follows.  
\begin{itemize}
    \item \textbf{Row Augmentation(RowAug)}: Select an existing row as a seed and modify its column data to simulate information of different individuals (e.g., changing names, adjusting numerical values). As augmented rows only serve to expand the dataset size without affecting the original problem's solvability, no additional validation of numerical rationality is required.
    \item \textbf{Column Augmentation(ColAug)}: A new column is added to the existing table. Since each row is constructed based on the protagonist of the mathematical problem, column augmentation enriches the description of the entities by adding information such as "height", "blood pressure", and other attributes.
    \item  \textbf{Order Shuffling(OrdShf)}: Randomly shuffle row or column sequences to increase the difficulty of data retrieval.
    \item \textbf{Information Modification(InfMod)}: This strategy affects solvability in two ways:
    \begin{itemize}
        \item \textbf{Missing Condition Modification}: Remove one or more key data points from seed rows (set as null), rendering the original problem unsolvable due to insufficient conditions.
        \item \textbf{Contradictory Condition Modification}: Modify implicit variables (values not explicitly stated but derivable from given conditions) to create conflicts with existing constraints, making the problem unsolvable due to logical contradictions.
    \end{itemize}
\end{itemize}

\newpage
\subsection{Experiment Details}

\subsubsection{Setup}
\label{app:ex setup}
\textbf{Models.} We evaluated four major categories of LLMs within \benchmark, including open-source general-purpose models (e.g., the Qwen series~\citep{yang2024qwen25}(including Qwen 3 and Qwen 2.5) and Llama3 series~\citep{grattafiori2024llama3}), open-source math-specialized models (DeepSeek-Math~\citep{shao2024deepseekmath} and Qwen-Math~\citep{yang2024qwen25math}), open-source table-specialized models (TableGPT~\citep{zha2023tablegptv2} and StructLM~\citep{zhuang2024structlm}), and proprietary API models (GPT-4~\citep{GPT4}, DeepSeek-v3~\citep{liu2024deepseek}, and GLM-4-plus~\citep{glm2024chatglm}).


\textbf{Formats.}
We evaluate three widely used ways of organizing tables: \textit{serialized format} \textit{Markdown format} and \textit{json format}. In the serialized format, each table row is converted into key-value pairs (e.g., \texttt{"name: Janet, Eggs\_per\_day: 16, Eggs\_eat\_morning: 4..."}). In the Markdown format, the table is presented using standard Markdown syntax, with the first row as column headers and subsequent rows listing values in order, using the \texttt{"|"} symbol as the column delimiter. In the \textit{json format}, Each row of the table will be organized into a format similar to JSON key-value pairs for the large model. Details prompts can be found in the appendix or the code section.

\textbf{Computing Resources.} We use NVIDIA A100 servers as our primary computing platform, along with a few additional machines equipped with RTX 4090 GPUs.

\subsubsection{Additional results}
\label{app:ex addi results}
\textbf{\Pipeline-generated data improves performance on other tabular reasoning datasets.} We evaluate on three other tabular math reasoning datasets on Qwen-2.5-7B model, namely TAT-QA~\citep{zhu2021tatqa}, FinQA~\citep{chen2021finqa}, and TabMWP~\citep{lu2022tabmwp}, which primarily test mathematical reasoning over tables. Under the same number of training steps, we further compare two settings: (i) training only on the target dataset’s official training set (\textit{Pure-finetune}), and (ii) training on a mixed dataset that combines the target dataset with data generated by \Pipeline(\textit{Mix-finetune}). As shown in Table~\ref{tab:train OOD}, the \textit{Mix-finetune} consistently outperforms the \textit{Pure-finetune} setting. The improvement is particularly pronounced on more complex tables, which highlights the versatility and generalization ability of our \Pipeline across diverse datasets.

\begin{table*}[t]
\centering
\caption{Performance of fine-tuned models on different tabular reasoning benchmarks}
\label{tab:train OOD}
\vspace{-0.5em}
\begin{tabular}{l|cccccc}
\toprule
\multirow{2}{*}{Setting} & \multicolumn{2}{c}{TabMWP} & \multicolumn{2}{c}{FinQA} & \multicolumn{2}{c}{TAT-QA} \\ 
\cmidrule(lr){2-3} \cmidrule(lr){4-5} \cmidrule(lr){6-7}
 & Average & Top10\% & Average & Top10\% & Average & Top10\% \\ \midrule
Baseline     & 74.09 & 74.07 & 57.72 & 48.00 & 37.34 & 30.77 \\
Pure-finetune & 91.60 & 91.90 & 74.80 & 71.19 & 67.38 & 51.28 \\
\midrule
Mix-finetune & 94.70 & 98.11 & 77.59 & 77.97 & 69.53 & 56.41 \\
\rowcolor{gray!15}
\multicolumn{1}{c|}{$\Delta$}      & {\color{red}$\uparrow$ 3.10} & {\color{red}$\uparrow$ 6.21} & {\color{red}$\uparrow$ 2.79} & {\color{red}$\uparrow$ 6.78} & {\color{red}$\uparrow$ 2.15} & {\color{red}$\uparrow$ 5.13} \\ 
\bottomrule
\end{tabular}
\end{table*}

\begin{figure*}[t]
    \centering
    \includegraphics[width=\linewidth]{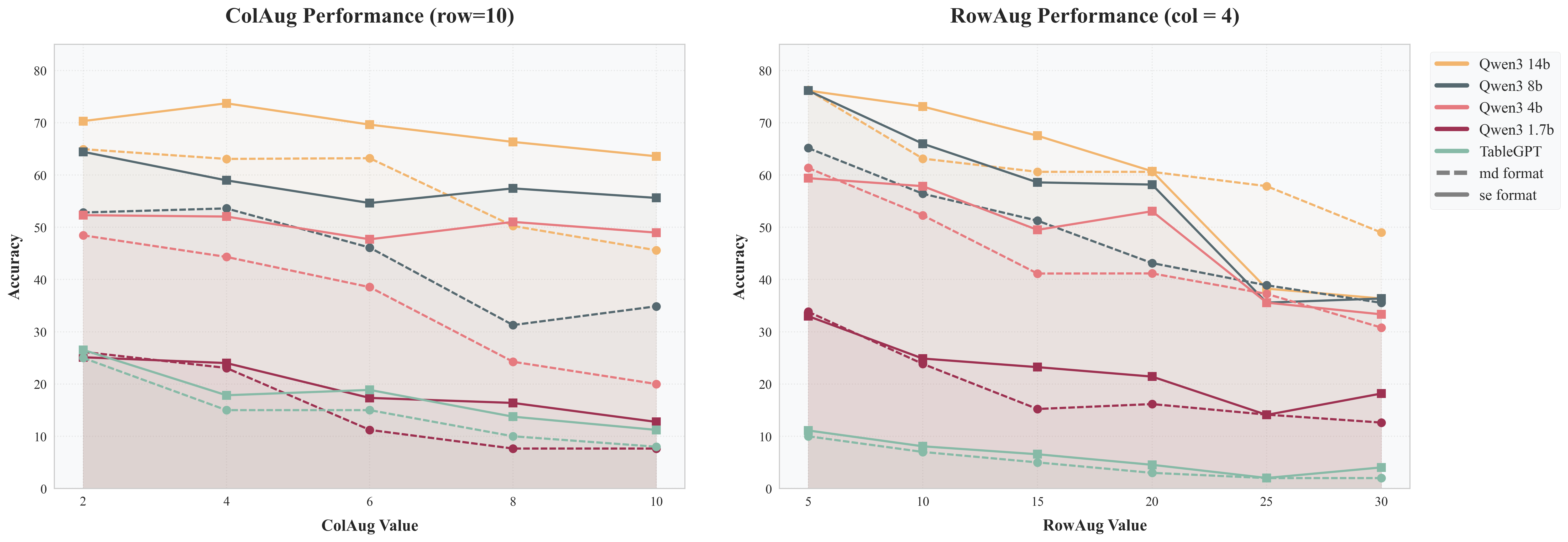}
    \caption{Relationship between model performance changes and table complexity}
    \label{fig: colaug/rowaug ablation}
\end{figure*}

\begin{table*}[t]
\centering
\vspace{-1em}
\caption{Comparison between performance on Direct trap and Hidden trap problems}
\label{tab: trap type}
\resizebox{0.9\textwidth}{!}{
\begin{tabular}{l|l|ccc|ccc}
\toprule
\midrule
                             &                       & \multicolumn{3}{c|}{Missing}                               & \multicolumn{3}{c}{Contra}                                \\ \cline{3-8} 
\multirow{-2}{*}{Model}      & \multirow{-2}{*}{Fmt} & Direct trap & Hidden trap & $\Delta$                          & Direct trap & Hidden trap & $\Delta$                          \\ \hline
                             & Se                    & 92.30       & 69.23       & \cellcolor[HTML]{EFEFEF}-23.07 & 28.68       & 28.57       & \cellcolor[HTML]{EFEFEF}-0.11 \\
\multirow{-2}{*}{Qwen3 14B}  & Md                    & 89.18       & 67.03       & \cellcolor[HTML]{EFEFEF}-22.15 & 21.47       & 28.21       & \cellcolor[HTML]{EFEFEF}6.74  \\
                             & Se                    & 93.79       & 70.97       & \cellcolor[HTML]{EFEFEF}-22.82 & 40.87       & 41.25       & \cellcolor[HTML]{EFEFEF}0.38  \\
\multirow{-2}{*}{Qwen3 8B}   & Md                    & 89.38       & 68.57       & \cellcolor[HTML]{EFEFEF}-20.81 & 31.47       & 34.54       & \cellcolor[HTML]{EFEFEF}3.07 \\
                             & Se                    & 91.72       & 68.67       & \cellcolor[HTML]{EFEFEF}-23.05 & 40.20       & 39.02       & \cellcolor[HTML]{EFEFEF}-1.18 \\
\multirow{-2}{*}{Qwen3 4B}   & Md                    & 88.19       & 69.75       & \cellcolor[HTML]{EFEFEF}-18.44 & 34.69       & 32.12       & \cellcolor[HTML]{EFEFEF}-2.57 \\
                             & Se                    & 50.17       & 37.33       & \cellcolor[HTML]{EFEFEF}-12.84 & 20.00       & 16.86       & \cellcolor[HTML]{EFEFEF}-3.14 \\
\multirow{-2}{*}{Qwen3 1.7B} & Md                    & 46.85       & 22.40       & \cellcolor[HTML]{EFEFEF}-24.45 & 16.72       & 14.40       & \cellcolor[HTML]{EFEFEF}-2.32 \\ 
\midrule
\bottomrule
\end{tabular}
}
\end{table*}

\textbf{Model performance degrades with increased retrieval difficulty.} First, we want to explore the relationship between performance degradation and table complexity. Through experiements in main text, we get an initial observation that model performance decreases monotonically as table complexity increases—from easy to hard levels. To further investigate the underlying mechanisms, we conducted a supplementary analysis based on two data augmentation strategies: \textit{ColAug} and \textit{RowAug}. 
We generated a series of augmented tables by fixing either the number of rows or columns and varying the other, to examine how model performance responds to changes in table structure.
As shown in Figure \ref{fig: colaug/rowaug ablation}, while the inference performance fluctuates as the number of columns (\textit{ColAug}) or rows (\textit{RowAug}) increases, a clear downward trend is evident. We attribute this degradation to the increased presence of irrelevant or distracting information, which raises the difficulty of information retrieval and subsequently impairs the model’s reasoning ability.

\textbf{Traps within the reasoning process make the model more prone to hallucinations.}
Which types of traps are more challenging for the model to detect? To explore this question, we introduce an additional set of experiments by incorporating two types of adversarial scenarios: Direct Missing (DM) and Direct Contra (DC). Unlike the trap questions involved in \benchmark, these traps are more explicit and thus easier to detect for humans.
In the DM questions, the table lacks the "name" attribute required by the question, which means the name of the target person mentioned in the question and their corresponding information are entirely absent from the table.
In the DC questions, the table contains two columns with the same header (i.e., duplicate column names) but with conflicting values. These conflicting entries can lead to different answers depending on which value is used. 
We present our experimental results in Table \ref{tab: trap type}. For missing traps, the model exhibits a significantly higher success rate in identifying direct traps compared to indirect ones. In contrast, performance on contra traps remains consistently poor across models, with some degree of variability. These results indicate that traps embedded within the reasoning process are inherently more difficult to detect.

\subsubsection{Detailed results}
We provide comprehensive experimental results here that are not given in the main text due to space limitations. Table~\ref{Table: main results appendix}~ is the complete version of Table~\ref{tab:main_results_left} and Table~\ref{tab:main_results_robust}. Table~\ref{tab:Retrieval}~ corresponds to Figure~\ref{fig: findings 2}~, which shows the performance comparison of table reasoning and single-step table retrieval. Table~\ref{tab:ColAug md row=10}~ ~\ref{tab:ColAug se row=10}~ ~\ref{tab:RowAug md}~ and ~\ref{tab:RowAug se}~ correspond to Figure~\ref{fig: colaug/rowaug ablation}, which shows the relationship between model performance changes and table complexity.

\begin{table*}[ht]
\caption{Main reults on \benchmark benchmark}
\label{Table: main results appendix}
\vspace{-1em}
\resizebox{\textwidth}{!}{
\begin{threeparttable}
\begin{tabular}{lcccccccccc}
\toprule
\midrule
\multicolumn{1}{l|}{}                                            & \multicolumn{1}{c|}{}                                 & \multicolumn{1}{c|}{}                      & \multicolumn{4}{c|}{Pure test}                                                    & \multicolumn{4}{c}{Robustness test}               \\ \cline{4-11} 
\multicolumn{1}{l|}{\multirow{-2}{*}{Dataset}}                   & \multicolumn{1}{c|}{\multirow{-2}{*}{GSM8k}}          & \multicolumn{1}{c|}{\multirow{-2}{*}{Fmt}} & Easy           & Medium         & Hard           & \multicolumn{1}{c|}{Avg}       & Well           & Contra         & Missing & Avg   \\ \hline
\multicolumn{11}{l}{\cellcolor[HTML]{FFCCC9}\textbf{Open source General Model}}                                                                                                                                                                                                                               \\ \hline
\multicolumn{1}{l|}{\multirow{2}{*}{Qwen3 14B}}        & \multicolumn{1}{l|}{\multirow{2}{*}{94.54}} & \multicolumn{1}{l|}{Se}                                        & 79.94                     & 70.94                       & 69.79 & \multicolumn{1}{l|}{73.55} & 58.55  & 28.57   & 69.23    & 54.15 \\
\multicolumn{1}{l|}{}                                  & \multicolumn{1}{l|}{}                       & \multicolumn{1}{l|}{Md}                                        & 77.87                     & 70.21                       & 61.59 & \multicolumn{1}{l|}{69.89} & 55.89  & 28.21   & 67.03    & 51.88 \\
\multicolumn{1}{l|}{\multirow{2}{*}{Qwen3 8B}}         & \multicolumn{1}{l|}{\multirow{2}{*}{93.30}} & \multicolumn{1}{l|}{Se}                                        & 75.17                     & 62.62                       & 56.99 & \multicolumn{1}{l|}{64.92} & 44.44  & 41.25   & 70.97    & 50.35 \\
\multicolumn{1}{l|}{}                                  & \multicolumn{1}{l|}{}                       & \multicolumn{1}{l|}{Md}                                        & 73.18                     & 55.63                       & 47.30 & \multicolumn{1}{l|}{58.70} & 36.71  & 34.54   & 68.57    & 44.07 \\
\multicolumn{1}{l|}{\multirow{2}{*}{Qwen3 4B}}         & \multicolumn{1}{l|}{\multirow{2}{*}{91.79}} & \multicolumn{1}{l|}{Se}                                        & 73.53                     & 56.85                       & 46.47 & \multicolumn{1}{l|}{58.95} & 39.63  & 39.02   & 68.67    & 46.77 \\
\multicolumn{1}{l|}{}                                  & \multicolumn{1}{l|}{}                       & \multicolumn{1}{l|}{Md}                                        & 71.57                     & 52.68                       & 41.73 & \multicolumn{1}{l|}{55.32} & 39.31  & 32.12   & 69.75    & 45.15 \\
\multicolumn{1}{l|}{\multirow{2}{*}{Qwen3 1.7B}}       & \multicolumn{1}{l|}{\multirow{2}{*}{81.25}} & \multicolumn{1}{l|}{Se}                                        & 54.40                     & 31.13                       &  19.20 & \multicolumn{1}{l|}{34.92}      &   26.81     &  16.86       &     45.71     &   29.03    \\
\multicolumn{1}{l|}{}                                  & \multicolumn{1}{l|}{}                       & \multicolumn{1}{l|}{Md}                                        & 50.76                     & 30.62                       & 18.84 & \multicolumn{1}{l|}{33.40}      &    33.33    &    14.40     &  37.33        &  29.56     \\
\multicolumn{1}{l|}{\multirow{2}{*}{Qwen2.5 14B}}      & \multicolumn{1}{l|}{\multirow{2}{*}{93.40}} & \multicolumn{1}{l|}{Se}                                        & 79.90                     & 68.06                       & 62.34 & \multicolumn{1}{l|}{70.10} & 59.40  & 6.40    & 22.40    & 36.90 \\
\multicolumn{1}{l|}{}                                  & \multicolumn{1}{l|}{}                       & \multicolumn{1}{l|}{Md}                                        & 79.21                     & 64.10                       & 49.09 & \multicolumn{1}{l|}{64.13} & 60.00  & 6.80    & 20.00    & 36.70 \\
\multicolumn{1}{l|}{\multirow{2}{*}{Qwen2.5 14B coder}} & \multicolumn{1}{l|}{\multirow{2}{*}{90.68}} & \multicolumn{1}{l|}{Se}                                        & 71.59                     & 60.38                       & 49.90 & \multicolumn{1}{l|}{60.62} & 47.40  & 30.00   & 52.40    & 44.30 \\
\multicolumn{1}{l|}{}                                  & \multicolumn{1}{l|}{}                       & \multicolumn{1}{l|}{Md}                                        & 72.63                     & 57.74                       & 45.61 & \multicolumn{1}{l|}{58.66} & 47.40  & 23.60   & 51.60    & 42.50 \\
\multicolumn{1}{l|}{\multirow{2}{*}{Qwen2.5 7B}}       & \multicolumn{1}{l|}{\multirow{2}{*}{82.86}} & \multicolumn{1}{l|}{Se}                                        & 35.56                     & 21.36                       & 19.39 & \multicolumn{1}{l|}{25.43} & 39.20  & 13.60   & 34.00    & 31.50 \\
\multicolumn{1}{l|}{}                                  & \multicolumn{1}{l|}{}                       & \multicolumn{1}{l|}{Md}                                        & 53.92                     & 34.45                       & 20.64 & \multicolumn{1}{l|}{36.33} & 37.40  & 16.00   & 34.80    & 31.40 \\
\multicolumn{1}{l|}{\multirow{2}{*}{Qwen2.5 7B coder}} & \multicolumn{1}{l|}{\multirow{2}{*}{84.71}} & \multicolumn{1}{l|}{Se}                                        & 62.35                     & 42.01                       & 29.79 & \multicolumn{1}{l|}{44.47} & 30.80  & 24.40   & 43.60    & 32.40 \\
\multicolumn{1}{l|}{}                                  & \multicolumn{1}{l|}{}                       & \multicolumn{1}{l|}{Md}                                        & 64.78                     & 42.13                       & 23.52 & \multicolumn{1}{l|}{43.47} & 33.40  & 20.80   & 34.00    & 30.40 \\
\multicolumn{1}{l|}{\multirow{2}{*}{Qwen2.5 3B}}       & \multicolumn{1}{l|}{\multirow{2}{*}{80.28}} & \multicolumn{1}{l|}{Se}                                        & 36.37                     & 22.71                      & 16.94  & \multicolumn{1}{l|}{25.34} & 2.20   & 84.00   & 91.20    & 44.90 \\
\multicolumn{1}{l|}{}                                  & \multicolumn{1}{l|}{}                       & \multicolumn{1}{l|}{Md}                                        & 39.74                     & 23.96                       & 15.68  & \multicolumn{1}{l|}{26.46} & 6.20   & 69.60   & 79.20    & 40.30 \\
\multicolumn{1}{l|}{\multirow{2}{*}{LLama3.1 8B}}      & \multicolumn{1}{l|}{\multirow{2}{*}{83.69}} & \multicolumn{1}{l|}{Se}                                        & 42.84                     & 34.93                       & 30.01 & \multicolumn{1}{l|}{35.92} & 29.00  & 6.40    & 8.40     & 18.20 \\
\multicolumn{1}{l|}{}                                  & \multicolumn{1}{l|}{}                       & \multicolumn{1}{l|}{Md}                                        & 48.61                     & 33.37                       & 32.15 & \multicolumn{1}{l|}{38.04} & 29.60  & 10.40   & 9.60     & 19.80 \\
\multicolumn{1}{l|}{\multirow{2}{*}{LLama3 8B}}        & \multicolumn{1}{l|}{\multirow{2}{*}{55.34}} & \multicolumn{1}{l|}{Se}                                        & 28.92                     & 15.22                       & 10.63 & \multicolumn{1}{l|}{18.25} & 12.80  & 30.80   & 37.20    & 23.40 \\
\multicolumn{1}{l|}{}                                  & \multicolumn{1}{l|}{}                       & \multicolumn{1}{l|}{Md}                                        & 36.30                     & 21.12                       & 20.68 & \multicolumn{1}{l|}{26.03} & 16.60  & 19.20   & 35.20    & 21.90 \\ \hline
\multicolumn{11}{l}{\cellcolor[HTML]{FFCCC9}\textbf{Open-Source Math Model}}                                                                                                                                                                                                                                  \\ \hline
\multicolumn{1}{l|}{\multirow{2}{*}{Qwen math 7B}}     & \multicolumn{1}{l|}{\multirow{2}{*}{95.45}} & \multicolumn{1}{l|}{Se}                                        & 53.69                     & 31.09                       & 14.59 & \multicolumn{1}{l|}{33.12} & 28.60  & 26.40   & 36.40    & 30.00 \\
\multicolumn{1}{l|}{}                                  & \multicolumn{1}{l|}{}                       & \multicolumn{1}{l|}{Md}                                        & 53.69                     & 30.37                       & 14.59 & \multicolumn{1}{l|}{32.88} & 27.00  & 20.80   & 48.93    & 30.93 \\
\multicolumn{1}{l|}{\multirow{2}{*}{DeepSeek math 7B}} & \multicolumn{1}{l|}{\multirow{2}{*}{80.13}} & \multicolumn{1}{l|}{Se}                                        & 13.93                     & 6.24                        & 3.96  & \multicolumn{1}{l|}{8.04}  & 2.60   & 50.40   & 51.20    & 26.70 \\
\multicolumn{1}{l|}{}                                  & \multicolumn{1}{l|}{}                       & \multicolumn{1}{l|}{Md}                                        & 12.81                     & 6.60                       & 2.04 & \multicolumn{1}{l|}{7.15} & 4.00   & 60.40   & 53.60    & 30.50  \\
\hline
\multicolumn{11}{|l|}{\cellcolor[HTML]{FFCCC9}\textbf{Open-Source Tabular Model}}                                                                                                                                                                                                                             \\ \hline
\multicolumn{1}{l|}{\multirow{2}{*}{TableGPT 7B}}      & \multicolumn{1}{l|}{\multirow{2}{*}{24.33}} & \multicolumn{1}{l|}{Se}                                        & 30.13                     & 18.86                       & 12.60 & \multicolumn{1}{l|}{20.53}                      & 26.20  & 26.80   & 44.80    & 31.00 \\
\multicolumn{1}{l|}{}                                  & \multicolumn{1}{l|}{}                       & \multicolumn{1}{l|}{Md}                                        & 30.60                     & 16.44                       & 17.64 &\multicolumn{1}{l|} {21.56  }                    & 30.60  & 23.20   & 46.40    & 32.70 \\
\multicolumn{1}{l|}{\multirow{2}{*}{StructLM 7B}}      & \multicolumn{1}{l|}{\multirow{2}{*}{32.97}} & \multicolumn{1}{l|}{Se}                                        & 13.74                     & 6.12                        & 3.24  &\multicolumn{1}{l|}{7.70 }                       & 7.20    & 0       & 0        & 3.60   \\
\multicolumn{1}{l|}{}                                  & \multicolumn{1}{l|}{}                       & \multicolumn{1}{l|}{Md}                                        & 14.78                     & 8.28                        & 4.44  & \multicolumn{1}{l|}{9.17  }                     & 9.60    & 0       & 0        & 4.80   \\ 
\hline
\multicolumn{11}{|l|}{\cellcolor[HTML]{FFCCC9}\textbf{Closed-Source API}}                                                                                                                                                                                                                                     \\ \hline
\multicolumn{1}{l|}{\multirow{2}{*}{DeepSeek v3}}      & \multicolumn{1}{l|}{\multirow{2}{*}{96.36}} & \multicolumn{1}{l|}{Se}                                        & 88.45                     & 87.27                       & 85.71 & \multicolumn{1}{l|}{87.14} & 68.60  & 68.40   & 85.20    & 72.70 \\
\multicolumn{1}{l|}{}                                  & \multicolumn{1}{l|}{}                       & \multicolumn{1}{l|}{Md}                                        & 88.63                     & 87.63                       & 85.83 & \multicolumn{1}{l|}{87.37} & 68.60  & 68.00   & 82.80    & 72.00 \\
\multicolumn{1}{l|}{\multirow{2}{*}{GLM-4-plus}}              & \multicolumn{1}{l|}{\multirow{2}{*}{95.07}} & \multicolumn{1}{l|}{Se}                                        & 83.37                     & 81.15                       & 79.83 & \multicolumn{1}{l|}{81.45} & 68.80  & 32.80   & 69.60    & 60.00 \\
\multicolumn{1}{l|}{}                                  & \multicolumn{1}{l|}{}                       & \multicolumn{1}{l|}{Md}                                        & 84.52                     & 81.03                       & 78.27 & \multicolumn{1}{l|}{81.27} & 71.40  & 27.60   & 65.60    & 59.00 \\
\multicolumn{1}{l|}{\multirow{2}{*}{GPT 4}}            & \multicolumn{1}{l|}{\multirow{2}{*}{94.46}} & \multicolumn{1}{l|}{Se}                                        & 83.97                     & 82.57                       & 77.41 & \multicolumn{1}{l|}{81.32} & 66.39  & 22.48   & 74.01    & 57.00 \\
\multicolumn{1}{l|}{}                                  & \multicolumn{1}{l|}{}                       & \multicolumn{1}{l|}{Md}                                        & 85.54                     & 78.42                       & 75.23 & \multicolumn{1}{l|}{79.73} & 64.25  & 21.11   & 80.20    & 57.80
 \\
\midrule
\bottomrule
\end{tabular}
\end{threeparttable}
}
\end{table*}

\begin{table*}[t]
\caption{Performance comparison under different reasoning settings across difficulty levels}
\label{tab:code comparation}
\centering
\small
\setlength{\tabcolsep}{4pt}
\begin{tabular}{lccccccccc}
\toprule
\multirow{3}{*}{\centering Model}
& \multicolumn{9}{c}{Difficulty Level} \\
\cmidrule(lr){2-10}
& \multicolumn{3}{c}{Easy}
& \multicolumn{3}{c}{Medium}
& \multicolumn{3}{c}{Hard} \\
\cmidrule(lr){2-4} \cmidrule(lr){5-7} \cmidrule(lr){8-10}
& Direct & Code+Hint & Code
& Direct & Code+Hint & Code
& Direct & Code+Hint & Code \\
\midrule
Qwen2.5 7B       
& 53.92 & 67.12 & 23.95
& 34.45 & 67.25 & 26.73
& 20.64 & 64.74 & 37.01 \\

Qwen2.5 7B Coder 
& 64.78 & 72.33 & 49.13
& 53.32 & 71.89 & 53.32
& 23.52 & 70.64 & 59.72 \\

Qwen2.5 14B      
& 79.21 & 72.70 & 64.89
& 64.10 & 71.52 & 66.88
& 49.09 & 72.40 & 65.50 \\

Qwen2.5 14B Coder
& 72.63 & 75.68 & 50.00
& 57.74 & 76.29 & 46.17
& 45.61 & 76.91 & 25.60 \\

Qwen3 14B        
& 77.87 & 72.46 & 59.93
& 70.21 & 70.51 & 55.71
& 61.59 & 71.14 & 60.81 \\

Qwen3 8B         
& 73.18 & 73.45 & 54.71
& 55.63 & 72.15 & 46.42
& 47.30 & 72.77 & 52.32 \\

Qwen3 4B         
& 71.57 & 69.73 & 33.87
& 52.68 & 67.50 & 35.01
& 41.73 & 67.63 & 38.77 \\

Qwen3 1.7B       
& 50.76 & 51.49 & 29.78
& 30.62 & 48.31 & 24.72
& 18.84 & 45.04 & 16.06 \\
\bottomrule
\end{tabular}
\vspace{-0.8em}
\end{table*}









\begin{table*}[ht]
\centering
\caption{Model Performance comparison of table reasoning and single-step table retrieval}
\label{tab:Retrieval}
\begin{tabular}{llcccc}
\toprule
Difficulty & Model & se-Retrival & se-Reason & md-Retrival & md-Reason \\
\midrule
\multirow{4}{*}{Easy}
  & Qwen3 14b  & 95.04 & 79.94 & 91.54 & 77.87 \\
  & Qwen3 8b   & 92.44 & 75.17 & 93.02 & 73.18 \\
  & Qwen3 4b   & 93.02 & 73.53 & 92.15 & 71.57 \\
  & Qwen3 1.7b & 81.10 & 54.40 & 85.71 & 50.76 \\
\midrule
\multirow{4}{*}{Medium}
  & Qwen3 14b  & 83.28 & 70.94 & 88.85 & 70.21 \\
  & Qwen3 8b   & 88.26 & 62.62 & 86.80 & 55.63 \\
  & Qwen3 4b   & 75.00 & 56.85 & 79.41 & 52.68 \\
  & Qwen3 1.7b & 63.82 & 31.13 & 69.50 & 30.62 \\
\midrule
\multirow{4}{*}{Hard}
  & Qwen3 14b  & 75.73 & 69.79 & 81.36 & 61.59 \\
  & Qwen3 8b   & 86.98 & 56.99 & 82.24 & 47.30 \\
  & Qwen3 4b   & 60.65 & 46.47 & 66.27 & 41.73 \\
  & Qwen3 1.7b & 58.28 & 19.25 & 56.80 & 18.84 \\
\bottomrule
\end{tabular}
\end{table*}

\begin{table*}[ht]
\centering
\caption{Model Performance Comparison with ColAug (md row=10)}
\label{tab:ColAug md row=10}
\begin{tabular}{lccccc}
\toprule
Model & ColAug2 & ColAug4 & ColAug6 & ColAug8 & ColAug10 \\
\midrule
Qwen3 14b   & 64.94 & 63.07 & 63.21 & 50.25 & 45.59 \\
Qwen3 8b    & 52.82 & 53.60 & 46.11 & 31.28 & 34.87 \\
Qwen3 4b    & 48.45 & 44.32 & 38.54 & 24.26 & 20.00 \\
Qwen3 1.7b  & 26.15 & 23.07 & 11.22 & 7.65  & 7.65  \\
LLaMA3 8b   & 33.67 & 31.12 & 31.63 & 23.97 & 18.87 \\
TableGPT    & 20.40 & 11.73 & 12.24 & 9.18  & 10.20 \\
\bottomrule
\end{tabular}
\end{table*}

\begin{table*}[ht]
\centering
\caption{Model Performance Comparison with ColAug (se row=10)}
\label{tab:ColAug se row=10}
\begin{tabular}{lccccc}
\toprule
Model & ColAug2 & ColAug4 & ColAug6 & ColAug8 & ColAug10 \\
\midrule
Qwen3 14b   & 70.31 & 73.71 & 69.63 & 66.32 & 63.58 \\
Qwen3 8b    & 64.43 & 58.97 & 54.63 & 57.43 & 55.61 \\
Qwen3 4b    & 52.30 & 52.04 & 47.69 & 51.02 & 48.97 \\
Qwen3 1.7b  & 25.12 & 24.01 & 17.34 & 16.38 & 12.75 \\
LLaMA3 8b   & 33.67 & 36.22 & 34.18 & 35.71 & 32.14 \\
TableGPT    & 26.53 & 17.85 & 18.87 & 13.75 & 11.22 \\
\bottomrule
\end{tabular}
\end{table*}

\begin{table*}[ht]
\centering
\caption{Model Performance Comparison with RowAug md}
\label{tab:RowAug md}
\begin{tabular}{lcccccc}
\toprule
Model & RowAug5 & RowAug10 & RowAug15 & RowAug20 & RowAug25 & RowAug30 \\
\midrule
Qwen3 14b   & 76.26 & 63.13 & 60.60 & 60.60 & 57.86 & 48.98 \\
Qwen3 8b    & 65.15 & 56.41 & 51.26 & 43.14 & 38.89 & 35.53 \\
Qwen3 4b    & 61.34 & 52.28 & 41.14 & 41.16 & 37.24 & 30.80 \\
Qwen3 1.7b  & 33.83 & 23.85 & 15.22 & 16.16 & 14.14 & 12.62 \\
LLaMA3 8b   & 40.40 & 38.88 & 36.36 & 36.68 & 28.28 & 20.20 \\
TableGPT    & 9.18  & 7.65  & 7.65  & 4.59  & 4.59  & 2.04  \\
\bottomrule
\end{tabular}
\end{table*}

\begin{table*}[ht]
\centering
\caption{Model Performance Comparison with RowAug se}
\label{tab:RowAug se}
\begin{tabular}{lcccccc}
\toprule
Model & RowAug5 & RowAug10 & RowAug15 & RowAug20 & RowAug25 & RowAug30 \\
\midrule
Qwen3 14b   & 76.14 & 73.09 & 67.51 & 60.71 & 38.25 & 36.36 \\
Qwen3 8b    & 76.14 & 65.98 & 58.58 & 58.16 & 35.57 & 36.36 \\
Qwen3 4b    & 59.39 & 57.86 & 49.49 & 53.06 & 35.57 & 33.33 \\
Qwen3 1.7b  & 32.99 & 24.87 & 23.23 & 21.42 & 14.09 & 18.18 \\
LLaMA3 8b   & 43.43 & 40.40 & 33.32 & 29.29 & 25.25 & 26.76 \\
TableGPT    & 11.11 & 8.08  & 6.56  & 4.54  & 2.02  & 4.04  \\
\bottomrule
\end{tabular}
\end{table*}

\clearpage

\begin{figure*}[thb]
    \begin{definedbox}[label=formalized_prompt]{Semantic Decoupling Prompt}
    \textbf{"system\_prompt":}
    \begin{quote}
    You are an experienced mathematician, and you are familiar with formal languages. I would like you to generate the formal form of a mathematical problem.

    You should express all logic in \textbf{SMT-LIB syntax}, using \textbf{prefix notation}. For example, multiplication should be written as `(* a b)` instead of `a * b`.

    HIGHLIGHT!!!:
    \textbf{All numbers appearing after 'assert' are written as floating point numbers.For example '2' is wrong and it should be replaced with '2.0'.}
    \end{quote}
    
    \vspace{0.5em}\hdashrule{\linewidth}{0.5pt}{2pt}\vspace{0.5em}
    
    \textbf{EXAMPLE INPUT:} 
    \begin{itemize}
        \item "problem": "Weng earns \$12 an hour for babysitting. Yesterday, she just did 50 minutes of babysitting. How much did she earn?"
    \end{itemize}

    \vspace{0.5em}\hdashrule{\linewidth}{0.5pt}{2pt}\vspace{0.5em}
    
    \textbf{EXAMPLE OUTPUT:}
    \begin{itemize}
        \item "problem": "Weng earns \$12 an hour for babysitting. Yesterday, she just did 50 minutes of babysitting. How much did she earn?",
        \item     "formal-problem":  
                "(declare-fun hourly\_rate () Int)\\
                (declare-fun minutes\_worked () Real)\\
                (declare-fun hours\_worked () Real)\\
                (declare-fun earnings () Real)\\
                (assert (= hourly\_rate 12.0))\\
                (assert (= minute\_worked 50.0))\\
                (assert (= minutes\_per\_hour 60.0))\\
                (assert (= hours\_worked (/ minutes\_per\_hour)))\\
                (assert (= earnings (* hourly\_rate hours\_worked)))\\
                (check-sat)\\
                (get-value (earnings))"
    \end{itemize}

    \vspace{0.5em}\hdashrule{\linewidth}{0.5pt}{2pt}\vspace{0.5em}

    \textbf{"user\_prompt:"}
    \begin{itemize}
        \item "problem": \{Question\}
    \end{itemize}
    
    \end{definedbox}
\end{figure*}

\begin{figure*}[thb]
    \begin{definedbox}[label=formalized_prompt]{Table Transformation Prompt}
    \small
    \textbf{"system\_prompt":}
    \begin{quote}
    The user will provide a problem and its formal representation. You need to convert the \textbf{explicitly assigned known data} of the problem into a tabular form.

The table should \textbf{only include variables that are directly assigned values in the problem} (e.g., via assertions like \texttt{(= variable value)}).

The table should include all variables that appear in the formal definition and their corresponding values: (``Given'' or ``Calculated'').

Please wrap the value of this variable and the method of obtaining it in a list like: \texttt{[5,"Given"]}

\textbf{Replace the variables} that appear in the table in the original problem with unknowns to generate a generalized problem (i.e., \texttt{table + generalization = original problem}). 

Set a \textbf{value range for each variable}, ensuring the ranges conform to common sense (they can be fixed values if appropriate).
    \end{quote}
    \hdashrule{\linewidth}{0.5pt}{2pt}
    
    \textbf{EXAMPLE INPUT:} 
    \begin{itemize}
    \item "problem": "Weng earns \$12 an hour for babysitting. Yesterday, she just did 50 minutes of babysitting. How much did she earn?",
    \item "formal\_problem": \\
                    "(declare-fun hourly\_rate () Real)\\
                      (declare-fun minutes\_worked () Int)\\
                      (declare-fun hours\_worked () Real)\\
                      (declare-fun earnings () Real)\\
                      (assert (= hourly\_rate 12.0))\\
                      (assert (= minutes\_worked 50))\\
                      (assert (= minutes\_per\_hour 60))\\
                      (assert (= hours\_worked (/ minutes\_worked minutes\_per\_hour)))\\
                      (assert (= earnings (* hourly\_rate hours\_worked)))\\
                      (check-sat)\\
                      (get-value (earnings))"
\end{itemize}
    \hdashrule{\linewidth}{0.5pt}{2pt}
    
    \textbf{EXAMPLE OUTPUT:}
    \begin{itemize}
    \item "problem": "Weng earns \$12 an hour for babysitting. Yesterday, she just did 50 minutes of babysitting. How much did she earn?",
    \item "table":  [
      "name": "Weng", \\
      "hourly\_rate": [12,"Given"], \\
      "minutes\_worked": [50,"Given"],\\
      "minutes\_per\_hour": [60,"Given"],\\
      "hours\_worked": [0.8333,"Calculated"],\\
      "earnings": [10,"Calculated"]
  ],
  \item "generalization": "Weng earns \$x an hour for babysitting. Yesterday, she just did t minutes of babysitting. How much did she earn?",
  \item "value\_ranges":\\
    "name": null,\\
    "hourly\_rate": 
      "min": 7.25,
      "max": 100,
      "unit": "dollars"\\
    "minutes\_worked": 
      "min": 10,
      "max": 1440,
      "unit": "minutes"\\
    "minutes\_per\_hour": 
      "min": 60,
      "max": 60,
      "unit": "minutes"
\end{itemize}
\hdashrule{\linewidth}{0.5pt}{2pt}

\textbf{"user\_prompt:"}
\begin{itemize}
    \item "problem": \{Question\}
    \item "formal\_problem": \{Formal\_problem\}
\end{itemize}
    \end{definedbox}
\end{figure*}
\clearpage

\subsection{Case Analysis}
\label{app:caseanalysis}
\subsubsection{Case Analysis of Hard Subset}
\begin{itemize}
    \item \textbf{Open source General model}: Take \textbf{Qwen14b} as an example (Figure~\ref{figure:Hard Qwen14b})

  \item \textbf{Open-Source Math model}: Take \textbf{Qwen math 7b} as an example (Figure~\ref{figure:Hard Qwen math 7b})
    
    \item \textbf{Open-Source Tabular model}: Take \textbf{TableGPT} as an example (Figure~\ref{figure:Hard TableGPT})

    \item \textbf{Closed source API}: Take \textbf{DeepSeek v3} as an example (Figure~\ref{figure:Hard DeepSeek v3})
\end{itemize}

\subsubsection{Case Analysis of Robust Subset}
\begin{itemize}
    \item \textbf{Open source General model}: Take \textbf{Qwen14b} as an example (Figure~\ref{figure:Robust Qwen14b})
    
  \item \textbf{Open-Source Math model}: Take \textbf{Qwen math 7b} as an example (Figure~\ref{figure:Robust Qwen math 7b})

    \item \textbf{Open-Source Tabular model}: Take \textbf{TableGPT} as an example (Figure~\ref{figure:Robust TableGPT})

    \item \textbf{Closed source API}: Take \textbf{DeepSeek v3} as an example (Figure~\ref{figure:Robust DeepSeek v3})
\end{itemize}

\begin{figure*}[thb]
    \caption{Hard --Qwen14b}
    \label{figure:Hard Qwen14b}
        \begin{definedbox}{Reason 1: Retrieval Failed}
        \textbf{Problem:} Eliza's rate per hour for the first x hours she works each week is \$y. She also receives an overtime pay of z times her regular hourly rate. If Eliza worked for t hours this week, how much are her earnings for this week?\\
        \begin{center}
        \scriptsize  
        \begin{tabular}{rrlrr>{\columncolor{red!15}}rrrr}
        \hline
        \textbf{OverRatio} & \textbf{OverHour} & \textbf{Name} & \textbf{BodyTemp} & \textbf{Age} & \textbf{RegularRate} & \textbf{Height} & \textbf{HeartRate} & \textbf{RegularHour} \\ 
        \hline
        \multicolumn{9}{c}{$\vdots$} \\
        1.2 & 5 & Eliza & 38 & 62 & 10.0 & 179 & 97 & 40 \\
        \multicolumn{9}{c}{$\vdots$} \\
        \hline
        \end{tabular}
        \end{center}
        
        \vspace{0.5em}\hdashrule{\linewidth}{0.5pt}{2pt}\vspace{0.5em}
        \normalsize
        
        \textbf{Model Response:} \textit{From the provided table, we can find that Eliza's regular hourly rate (y) is \$10.0, and her overtime multiplier (z) is 1.5. \textcolor{red}{The number of regular working hours before overtime starts is not directly given} but implied to be within the total hours worked as anything beyond that is considered overtime. ......}
        \end{definedbox}

        \begin{definedbox}{Reason 2: Reasoning Failed}
        \textbf{Problem:} Charlie wants to sell beeswax candles.  For every pound of beeswax, he can make x tapered candles.  One pound of beeswax and the wicks cost \$y in supplies.   If he sells each candle for \$z each, what is his net profit if he makes and sells t candles?\\
        \begin{center}
        \scriptsize  
        \begin{tabular}{rrlrrrrrr}
        \hline
        \textbf{HeartRate} & \textbf{BodyTemp} & \textbf{Name} & \textbf{Age} & \textbf{CandlesMade} & \textbf{cost/pound} & \textbf{Height} & \textbf{price} & \textbf{candles/pound} \\ 
        \hline
        \multicolumn{9}{c}{$\vdots$} \\
        83 & 36 & Charlie & 74 & 20 & 10.0 & 199 & 2.0 & 10 \\
        \multicolumn{9}{c}{$\vdots$} \\
        \hline
        \end{tabular}
        \end{center}
        
        \vspace{0.5em}\hdashrule{\linewidth}{0.5pt}{2pt}\vspace{0.5em}
        \normalsize
        
        \textbf{Model Response:} \textit{......we find that Charlie's specific information is: Cost\_Per\_Pound (y): \$10.0 Selling\_Price (z):\$2.0 Candles\_Per\_Pound (x): 10. However, there seems to be an inconsistency as the selling price per candle (z) should ideally be higher than the cost of producing one candle to ensure a profit. \textcolor{red}{Typically, the selling price (z) would be higher than the cost per unit, but here, it appears to be lower which would imply a loss rather than profit}......}
        \end{definedbox}
       
    \end{figure*}

\begin{figure*}[thb]
\caption{Hard --Qwen math 7b}
\label{figure:Hard Qwen math 7b}
    \begin{definedbox}{Reason 1: Retrieval Error}
    \textbf{Problem:} Janet’s ducks lay x eggs per day. She eats y for breakfast every morning and bakes muffins for her friends every day with z. She sells the remainder at the farmers' market daily for \$w per fresh duck egg. How much in dollars does she make every day at the farmers' market?\\
    \begin{center}
    \scriptsize  
    \begin{tabular}{rr>{\columncolor{red!15}}rrrrlrr}
    \hline
    \textbf{Age} & \textbf{HeartRate} & \textbf{eggs/day} & \textbf{price/egg} & \textbf{BodyTemp} & \textbf{eaten} & \textbf{name} & \textbf{for\_muffins} & \textbf{SleepHours} \\ 
    \hline
    \multicolumn{9}{c}{$\vdots$} \\
    25 & 72 & 16 & 2 & 39 & 3 & Janet & 4 & 5 \\
    \multicolumn{9}{c}{$\vdots$} \\
    \hline
    \end{tabular}
    \end{center}
    
    \vspace{0.5em}\hdashrule{\linewidth}{0.5pt}{2pt}\vspace{0.5em}
    \normalsize
    
    \textbf{Model Response:} \textit{......For Janet: \textcolor{red}{Eggs per day: 94}  ,Eggs eaten: 3,Eggs for muffins: 4. First, calculate the number of eggs Janet has left after eating and making muffins: 94 - 3 - 4 = 87 ......}
    \end{definedbox}
\end{figure*}

\begin{figure*}[thb]
\caption{Hard --TableGPT}
\label{figure:Hard TableGPT}
    \begin{definedbox}{Reason 1: No Ability to Solve}
    \textbf{Problem:} Kylar went to the store to buy glasses for his new apartment. One glass costs \$x, but every second glass costs only y\% of the price. Kylar wants to buy n glasses. How much does he need to pay for them?\\
    \begin{center}
    \scriptsize  
    \begin{tabular}{rrrlrrrr}
    \hline
    \textbf{price/glass} & \textbf{glasses} & \textbf{Weight} & \textbf{name} & \textbf{HeartRate} & \textbf{discount} & \textbf{BodyTemp} & \textbf{Height} \\ 
    \hline
    \multicolumn{8}{c}{$\vdots$} \\
    5.0 & 16 & 98 & Kylar & 98 & 0.6 & 37 & 78 \\
    \multicolumn{8}{c}{$\vdots$} \\
    \hline
    \end{tabular}
    \end{center}
    
    \vspace{0.5em}\hdashrule{\linewidth}{0.5pt}{2pt}\vspace{0.5em}
    \normalsize
    
    \textbf{Model Response:} \textit{This problem is unsolvable.}
    \end{definedbox}

    \begin{definedbox}{Reason 2: Retrieval Failed}
    \textbf{Problem:} Mishka bought x pairs of shorts, y pairs of pants, and z pairs of shoes. One pair of shorts costs \$a. One pair of pants costs \$b and one pair of shoes costs \$c. How many dollars did Mishka spend on all the clothing items?\\
    \begin{center}
    \scriptsize  
    \begin{tabular}{>{\columncolor{red!15}}rrrrr>{\columncolor{red!15}}rrrlr>{\columncolor{red!15}}r}
    \hline
    \textbf{pants} & \textbf{\makecell{pants\\price}} & \textbf{SleepHours} & \textbf{\makecell{shorts\\price}} & \textbf{\makecell{shoes\\price}} & \textbf{shorts} & \textbf{Age} & \textbf{Weight} & \textbf{name} & \textbf{Height} & \textbf{shoes}\\ 
    \hline
    \multicolumn{11}{c}{$\vdots$} \\
    3 & 22.5 & 8 & 16.5 & 42.0 & 3 & 21 & 53 & Mishka & 198 & 3\\
    \multicolumn{11}{c}{$\vdots$} \\
    \hline
    \end{tabular}
    \end{center}
    
    \vspace{0.5em}\hdashrule{\linewidth}{0.5pt}{2pt}\vspace{0.5em}
    \normalsize
    
    \textbf{Model Response:} \textit{......The table provides information about various individuals and their clothing costs, but \textcolor{red}{it does not directly provide the values for x, y, and z} (the number of pairs of shorts, pants, and shoes Mishka bought)......}
    \end{definedbox}
\end{figure*}

\begin{figure*}[thb]
    \caption{Hard --DeepSeek v3}
    \label{figure:Hard DeepSeek v3}
        \begin{definedbox}{Reason 1: Misunderstanding the Question}
        \textbf{Problem:} John plans to sell all his toys and use the money to buy video games. He has x lego sets and he sells them for \$y each. He ends up buying z video games for \$w each and has \$v left. How many lego sets does he still have?\\
        \begin{center}
        \scriptsize  
        \begin{tabular}{lrrrrrrrrr}
        \hline
        \textbf{name} & \textbf{\makecell{lego\\price}} & \textbf{\makecell{Sleep\\Hours}} & \textbf{\makecell{Body\\Temp}} & \textbf{\makecell{videogame\\price}} & \textbf{\makecell{video\\games}} & \textbf{\makecell{lego\\sets}} & \textbf{\makecell{Heart\\Rate}} & \textbf{\makecell{money\\left}} & \textbf{Height}\\ 
        \hline
        \multicolumn{10}{c}{$\vdots$} \\
        John & 15 & 8 & 40 & 20 & 8 & 13 & 70 & 5 & 159 \\
        \multicolumn{10}{c}{$\vdots$} \\
        \hline
        \end{tabular}
        \end{center}
        
        \vspace{0.5em}\hdashrule{\linewidth}{0.5pt}{2pt}\vspace{0.5em}
        \normalsize
        
        \textbf{Model Response:} \textit{......\textcolor{red}{Since John plans to sell all his toys, he sells all 13 lego sets. Therefore, he has 0 lego sets left}.}
        \end{definedbox}

        \begin{definedbox}{Reason 2: Reasoning Error}
        \textbf{Problem:} Sasha notices that prices for lumber have gone up x\% in the last few months after she bought some lumber. Since she has leftovers, she decides to sell them. She has leftover s small boards that cost her \$c each. She also has l large boards she bought for \$d each. If she sells them all, how much profit does she make?\\
        \begin{center}
        \scriptsize  
        \begin{tabular}{rrrlrrrrr>{\columncolor{red!15}}r}
        \hline
        \textbf{\makecell{small\\boards}} & \textbf{\makecell{cost\\large\\boards}} & \textbf{BodyTemp} & \textbf{name} & \textbf{\makecell{cost\\small\\boards}} & \textbf{Age} & \textbf{\makecell{large\\boards}} & \textbf{Height} & \textbf{Weight} & \textbf{\makecell{percentage\\increase}}\\ 
        \hline
        \multicolumn{10}{c}{$\vdots$} \\
        10 & 16.0 & 39 & Sasha & 10 & 37 & 5 & 150 & 51 & 0.5 \\
        \multicolumn{10}{c}{$\vdots$} \\
        \hline
        \end{tabular}
        \end{center}
        
        \vspace{0.5em}\hdashrule{\linewidth}{0.5pt}{2pt}\vspace{0.5em}
        \normalsize
        
        \textbf{Model Response:} \textit{......Selling price for small boards: ( c times (1 + x/100) = \textcolor{red}{ 10.0 times (1 + 0.5/100) = 10.0 times 1.005 = 10.05}.}
        \end{definedbox}
    \end{figure*}

\begin{figure*}[thb]
\caption{Robust --Qwen14b}
\label{figure:Robust Qwen14b}
    \begin{definedbox}{Reason 1: Retrieval Failed}
    \textbf{Problem:} Jordan has x children who wear diapers. Each child requires y diaper changes per day. Jordan's wife changes half of the diapers. How many diapers does Jordan change per day?\\
    \begin{center}
    \scriptsize  
    \begin{tabular}{ccl}
    \hline
    \textbf{children} & \textbf{diaper\_changes\_per\_child} & \textbf{name} \\ 
    \hline
    \multicolumn{3}{c}{$\vdots$} \\
    2 & 5 & Jordan \\
    \multicolumn{3}{c}{$\vdots$} \\
    \hline
    \end{tabular}
    \end{center}
    
    \vspace{0.5em}\hdashrule{\linewidth}{0.5pt}{2pt}\vspace{0.5em}
    \normalsize
    
    \textbf{Model Response:} \textit{........we see various entries for different names and their respective children and diaper change requirements. From the table, \textcolor{red}{the only mention of Jordan is as the name of a child with 2 children needing 5 diaper changes per day. But since the question implies that Jordan (as a parent) has x children, we need to clarify which set of children belong to Jordan as a parent. }Since this information isn't clearly specified in the question or table, it's not possible to determine exactly how many children Jordan has as a parent or their specific diaper needs.. ......}
    \end{definedbox}

    \begin{definedbox}{Reason 2: Fail to Identify Contradictions}
    \textbf{Problem:} Bud makes homemade macaroni and cheese once a week.  The pasta costs \$x a box, and he spends \$y on cheddar cheese and \textcolor{red}{twice that amount for the gruyere cheese}.  How much money does Bud spend on making macaroni and cheese in one year?\\
    \begin{center}
    \scriptsize  
    \begin{tabular}{>{\columncolor{red!15}}rrrrrl}
    \hline
    \textbf{\makecell{cost\\gruyere\\cheese}} & \textbf{\makecell{weeks\\per\\year}} & \textbf{\makecell{total\\cost\\week}} & \textbf{\makecell{cost\\cheddar\\cheese}} & \textbf{\makecell{cost\\pasta}} & \textbf{name} \\ 
    \hline
    \multicolumn{6}{c}{$\vdots$} \\
    10.0 & 52 & 5.0 & 3.0 & 1.0 & Bud \\
    \multicolumn{6}{c}{$\vdots$} \\
    \hline
    \end{tabular}
    \end{center}
    
    \vspace{0.5em}\hdashrule{\linewidth}{0.5pt}{2pt}\vspace{0.5em}
    \normalsize
    
    \textbf{Model Response:} \textit{........The cost of cheddar cheese (y) = \textcolor{red}{3.0}.  The cost of gruyere cheese =  \textcolor{red}{10.0}. \textcolor{red}{Since the cost of gruyere cheese is twice the cost of cheddar cheese, this aligns with the provided values.} The weekly cost for Bud is thus:Total weekly cost = x + y + 2y = 1.0 + 3.0 + 2(3.0) = 1.0 + 3.0 + 6.0 = 10.0......}
    \end{definedbox}
\end{figure*}

\begin{figure*}[thb]
\caption{Robust --Qwen math 7b}
\label{figure:Robust Qwen math 7b}
    \begin{definedbox}{Reason 1: Contradictions Identified but Wrong Answer}
    \textbf{Problem:} Andrew plans a road trip from New Jersey to Rochester. It takes x days to travel by bus and \textcolor{red}{half as many days} to travel by car from New Jersey to Rochester. How many days will he travel if he plans to go to Rochester by bus and return from Rochester by car?\\
    \begin{center}
    \scriptsize  
    \begin{tabular}{lr>{\columncolor{red!15}}r}
    \hline
    \textbf{name} & \textbf{bus\_days} & \textbf{car\_days}\\ 
    \hline
    \multicolumn{3}{c}{$\vdots$} \\
    Andrew & 6 & 2 \\
    \multicolumn{3}{c}{$\vdots$} \\
    \hline
    \end{tabular}
    \end{center}
    
    \vspace{0.5em}\hdashrule{\linewidth}{0.5pt}{2pt}\vspace{0.5em}
    \normalsize
    
    \textbf{Model Response:} \textit{........From the table, we can see that Andrew has a bus days value of 6 and a car days value of 2. The problem states that it takes half as many days to travel by car as it does to travel by bus. However, the table provides a different value for car days, which is 2. \textcolor{red}{This suggests that the table may have an error} or that the problem may have an additional condition that we need to consider.Given the information in the table, we can use the bus days value to calculate the total travel time. The total travel time is the sum of the bus days and the car days.\textcolor{red}{Total travel time = Bus days + Car days = 6 + 2 = 8}. Therefore, the total number of days Andrew will travel is 8......}
    \end{definedbox}

    \begin{definedbox}{Reason 2: Missing Identified but Wrong Answer}
    \textbf{Problem:} Aleksandra went to a restaurant for dinner. She ordered some soup, a bagel, and a piece of cake. The bagel cost \$x, and the soup 25\% more. The cake is only half of the price of the bagel. How much did Aleksandra need to pay for the dinner she ordered?\\
    \begin{center}
    \scriptsize  
    \begin{tabular}{rl}
    \hline
    \textbf{bagel\_cost} & \textbf{name}  \\ 
    \hline
    3.2 & Alexander \\
    15.8 & Aiden \\
    10.7 & Sophia \\
    \multicolumn{2}{c}{$\vdots$} \\
    \textcolor{red}{null} & Aleksandra \\
    \multicolumn{2}{c}{$\vdots$} \\
    \hline
    \end{tabular}
    \end{center}
    
    \vspace{0.5em}\hdashrule{\linewidth}{0.5pt}{2pt}\vspace{0.5em}
    \normalsize
    
    \textbf{Model Response:} \textit{......we are interested in the cost of the bagel ordered by Aleksandra, which is not specified in the table. However, we can assume that the table is meant to provide a reference for the cost of the bagel, and we can \textcolor{red}{use the average cost of the bagel from the table to find a reasonable estimate}.......}
    \end{definedbox}
\end{figure*}

\begin{figure*}[thb]
\caption{Robust --TableGPT}
\label{figure:Robust TableGPT}
     \begin{definedbox}{Reason 1: Fail to Identify Contradiction}
    \textbf{Problem:} Andrew plans a road trip from New Jersey to Rochester. It takes x days to travel by bus and \textcolor{red}{half as many days} to travel by car from New Jersey to Rochester. How many days will he travel if he plans to go to Rochester by bus and return from Rochester by car?\\
    \begin{center}
    \scriptsize  
    \begin{tabular}{lr>{\columncolor{red!15}}r}
    \hline
    \textbf{name} & \textbf{bus\_days} & \textbf{car\_days}\\ 
    \hline
    \multicolumn{3}{c}{$\vdots$} \\
    Andrew & 6 & 2 \\
    \multicolumn{3}{c}{$\vdots$} \\
    \hline
    \end{tabular}
    \end{center}
    
    \vspace{0.5em}\hdashrule{\linewidth}{0.5pt}{2pt}\vspace{0.5em}
    \normalsize
    
    \textbf{Model Response:} \textit{........The table provides the bus and car days for several individuals, including Andrew.  \textcolor{red}{From the table: Andrew's bus days: 6 , Andrew's car days: 2.} If Andrew plans to go to Rochester by bus and return from Rochester by car, the total number of days he will travel is the sum of his bus days and car days.  Total days = Bus days + Car days = 6 + 2 = 8......}
    \end{definedbox}

    \begin{definedbox}{Reason 2: Regard Missing Info as 0}
    \textbf{Problem:} Samwell owns an appliances store. For this week, one-fourth of their sales are smart TVs, one-eighth are analog TVs, and the rest are OLED TVs. If they were able to sell a total of x TVs, how many OLED TVs were they able to sell?\\
    \begin{center}
    \scriptsize  
    \begin{tabular}{rl}
    \hline
    \textbf{total\_tvs} & \textbf{name}  \\ 
    \hline
    708 & Sebastian \\
    503 & Sophia \\
    935 & Emily \\
    \multicolumn{2}{c}{$\vdots$} \\
    \textcolor{red}{null} & Aleksandra \\
    \multicolumn{2}{c}{$\vdots$} \\
    \hline
    \end{tabular}
    \end{center}
    
    \vspace{0.5em}\hdashrule{\linewidth}{0.5pt}{2pt}\vspace{0.5em}
    \normalsize
    
    \textbf{Model Response:} \textit{......\textcolor{red}{According to the table, Samwell's total TV sales are listed as 0 (None)}. Since Samwell's total TV sales are 0, we can calculate the number of OLED TVs sold by using the given fractions for smart TVs and analog TVs.One-fourth of the sales are smart TVs: 1/4 times 0 = 0. One-eighth of the sales are analog TVs:1/8 times 0 = 0. The rest of the sales are OLED TVs: 0 - 0 - 0 = 0.......}
    \end{definedbox}
\end{figure*}

\begin{figure*}[thb]
    \caption{Robust --DeepSeek v3}
    \label{figure:Robust DeepSeek v3}
         \begin{definedbox}{Reason 1: Contradictions Identified but Reasoning Error}
        \textbf{Problem:}\\
        \begin{center}
        \scriptsize  
        \begin{tabular}{rl>{\columncolor{red!15}}rr}
        \hline
        \textbf{first\_hour\_cost} & \textbf{name} & \textbf{subsequent\_hour\_cost} & \textbf{total\_hours} \\ 
        \hline
        \multicolumn{4}{c}{$\vdots$} \\
        15 & James & 35 & 4 \\
        \multicolumn{4}{c}{$\vdots$} \\
        \hline
        \end{tabular}
        \end{center}
        
        \vspace{0.5em}\hdashrule{\linewidth}{0.5pt}{2pt}\vspace{0.5em}
        \normalsize
        
        \textbf{Model Response:} \textit{......we find the row corresponding to James: First\_Hour\_Cost (x): 15, Subsequent\_Hour\_Cost:35 \textbf{(Note: This is not directly used in the calculation since the problem states each subsequent hour is twice the previous hour's cost, not a fixed subsequent cost)} Total\_Hours: 4. First paid hour:15. \textcolor{red}{Second paid hour:2 × 15 = 30. Third paid hour:2 × 30 = 60.} Total Cost = 15+ 3 + 60 = 105...... (The correct answer should be 15+30+30=75) }
        \end{definedbox}

        \begin{definedbox}{Reason 2: Regard Missing Info as 0}
        \textbf{Problem:} Elaina is holding the final concert in her tour. To celebrate her final concert, she makes the concert twice as long as her usual concerts. At the end of the concert, she also performs a t-minute encore. If the runtime of this final concert is f minutes then how long, in minutes, do her usual concerts run for?\\
        \begin{center}
        \scriptsize  
        \begin{tabular}{lrr}
        \hline
        \textbf{name} & \textbf{final\_concert\_length} & \textbf{encore\_length} \\ 
        \hline
        James & 146 & 22\\
        Ella & 115 & 5\\
        Matthew & 92 & 24\\
        \multicolumn{3}{c}{$\vdots$} \\
        Elaina & 65 & \textcolor{red}{null} \\
        \multicolumn{3}{c}{$\vdots$} \\
        \hline
        \end{tabular}
        \end{center}
        
        \vspace{0.5em}\hdashrule{\linewidth}{0.5pt}{2pt}\vspace{0.5em}
        \normalsize
        
        \textbf{Model Response:} \textit{......From the table, we have the following data for Elaina: Final\_Concert\_Length (f): 65. \textcolor{red}{Encore\_Length (t): None (which we can interpret as 0 minutes, since no encore is performed)} According to the problem, the final concert is twice as long as the usual concert, thus x = 65/2 = 32.5.......}
        \end{definedbox}
    \end{figure*}

\end{document}